\documentclass[10pt,twocolumn,letterpaper]{article}

\usepackage[pagenumbers]{cvpr} 

\graphicspath{{figures/}}

\usepackage{pifont}
\usepackage{booktabs}
\usepackage{multirow}
\usepackage{rotating}
\usepackage{algorithm}
\usepackage{algpseudocode}
\usepackage{capt-of}
\usepackage{url}
\usepackage{mathtools}
\usepackage{subcaption}
\usepackage{colortbl}
\usepackage{nicematrix}

\newcommand{\Tref}[1]{Table~\ref{#1}}

\newcommand{\Fref}[1]{Fig.~\ref{#1}}

\definecolor{cvprblue}{rgb}{0.21,0.49,0.74}
\usepackage[pagebackref,breaklinks,colorlinks,allcolors=cvprblue]{hyperref}

\def\paperID{*****} 
\def\confName{CVPR}
\def\confYear{2026}

\title{Causal Motion Diffusion Models for Autoregressive Motion Generation}

\author{
Qing Yu\textsuperscript{1}\quad
Akihisa Watanabe\textsuperscript{2}\quad
Kent Fujiwara\textsuperscript{1}\\
\textsuperscript{1}LY Corporation\qquad
\textsuperscript{2}Waseda University\\
{\tt\small \{yu.qing,~kent.fujiwara\}@lycorp.co.jp, akihisa@ruri.waseda.jp}
}

\begin{document}
\maketitle
\begin{abstract}
Recent advances in motion diffusion models have substantially improved the realism of human motion synthesis. However, existing approaches either rely on full-sequence diffusion models with bidirectional generation, which limits temporal causality and real-time applicability, or autoregressive models that suffer from instability and cumulative errors. In this work, we present Causal Motion Diffusion Models (CMDM), a unified framework for autoregressive motion generation based on a causal diffusion transformer that operates in a semantically aligned latent space. CMDM builds upon a Motion-Language-Aligned Causal VAE (MAC-VAE), which encodes motion sequences into temporally causal latent representations. On top of this latent representation, an autoregressive diffusion transformer is trained using causal diffusion forcing to perform temporally ordered denoising across motion frames. To achieve fast inference, we introduce a frame-wise sampling schedule with causal uncertainty, where each subsequent frame is predicted from partially denoised previous frames. The resulting framework supports high-quality text-to-motion generation, streaming synthesis, and long-horizon motion generation at interactive rates. Experiments on HumanML3D and SnapMoGen demonstrate that CMDM outperforms existing diffusion and autoregressive models in both semantic fidelity and temporal smoothness, while substantially reducing inference latency.
\end{abstract}    
\section{Introduction}
\label{sec:intro}

Synthesizing realistic human motion conditioned on natural language remains a fundamental problem in computer vision and graphics. A successful text-to-motion generation model should not only synthesize spatially accurate body movements but also maintain temporal coherence across long sequences. Recent progress in motion diffusion models~\cite{mdm2022human,zhang2022motiondiffuse,chen2023executing,dai2024motionlcm} has led to significant improvements in motion quality and diversity, benefiting from the strong generative capacity of diffusion-based frameworks~\cite{ho2020denoising,dhariwal2021diffusion}. However, most existing diffusion models rely on bidirectional denoising over the entire sequence, which inherently breaks temporal causality and prevents online generation.

\begin{figure}[t]
    \centering
    \includegraphics[width=1\linewidth]{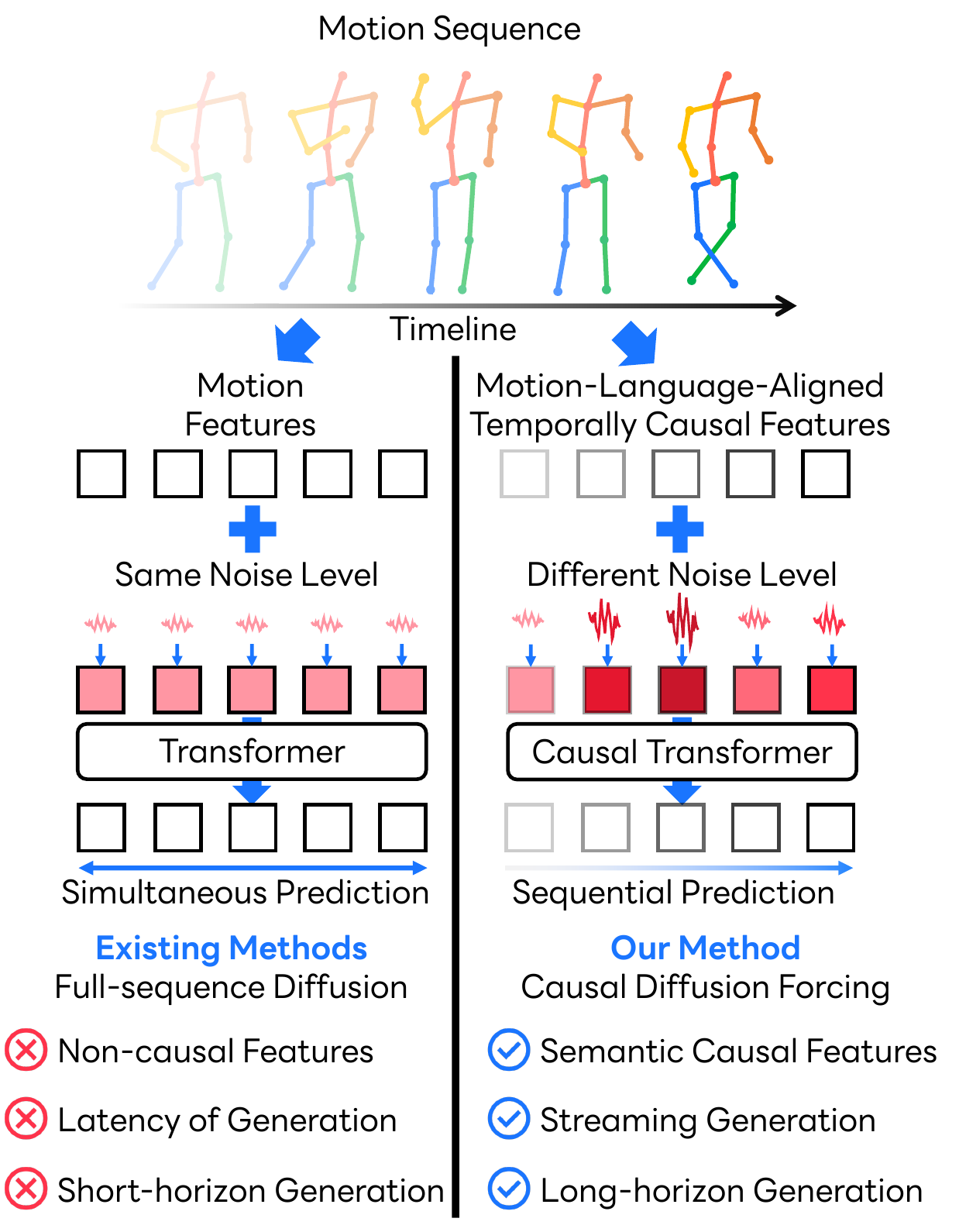}
    \vspace{-15pt}
    \caption{Overview of the existing methods and the proposed method. Existing diffusion-based methods (left) perform full-sequence denoising using the same noise level across all frames. In contrast, our proposed CMDM (right) introduces a causal diffusion forcing mechanism that operates on semantic causal latent features with frame-wise noise levels.}
    \vspace{-10pt}
    \label{fig:setting}
\end{figure}

Autoregressive models~\cite{zhang2023t2m,meng2024rethinking,zhao2024dartcontrol,xiao2025motionstreamer} offer an alternative by predicting future frames from past ones, ensuring causal consistency and supporting online motion generation. Yet, their sequential dependency often leads to error accumulation, making long-horizon synthesis unstable and inefficient. The key challenge lies in achieving temporally ordered, high-quality motion generation with both the fidelity of diffusion models and the causal structure of autoregressive transformers.

To address these challenges, we propose \textbf{C}ausal \textbf{M}otion \textbf{D}iffusion \textbf{M}odels (\textbf{CMDM}), a unified framework that integrates causal diffusion and autoregressive modeling within a semantically aligned latent space as shown in~\Fref{fig:setting}. CMDM is built upon our Motion-Language-Models-Aligned Causal Variational Autoencoder (MAC-VAE), which encodes human motion into temporally causal latent representations guided by motion-language pretraining~\cite{radford2021learning,petrovich2023tmr,yu2024exploring}. This foundation enables CMDM to operate in a compact and semantically meaningful latent space, preserving alignment between linguistic semantics and motion dynamics. On top of MAC-VAE, we design a Causal Diffusion Transformer (Causal-DiT) that performs diffusion denoising in an autoregressive manner. Unlike conventional diffusion models that jointly process all frames, Causal-DiT applies causal self-attention to ensure each frame depends only on preceding frames. This design enforces strict temporal ordering, allowing streaming motion generation.

To accelerate inference, we introduce a frame-wise sampling schedule with causal uncertainty, which allows each frame to be progressively refined from partially denoised preceding frames rather than requiring a fully autoregressive denoising step. Inspired by Diffusion Forcing~\cite{chen2024diffusion}, originally designed for next-token prediction, during training we perturb each frame with independent noise levels while maintaining causal dependencies across time, enabling the model to learn temporally consistent denoising transitions. During sampling, the model iteratively predicts the subsequent frames based on previously denoised frames with varying noise levels, gradually reducing uncertainty in a causal order. This hierarchical denoising process significantly reduces inference steps, achieving efficient and temporally coherent motion generation.

CMDM unifies the stability and realism of diffusion models with the causality and efficiency of autoregressive architectures. The framework enables high-fidelity text-to-motion generation, fast inference, and long-horizon synthesis within a unified causal formulation. Extensive evaluations on HumanML3D and SnapMoGen demonstrate that CMDM consistently outperforms state-of-the-art diffusion and autoregressive models in both semantic fidelity and temporal smoothness, while reducing inference latency by an order of magnitude.

Our main contributions are summarized as follows:
\begin{itemize}
\setlength\itemsep{0pt}
\item \textbf{Causal motion diffusion framework.} We propose CMDM, the first motion diffusion framework that unifies causal autoregression and diffusion denoising within a motion–language–aligned latent space.

\item \textbf{Semantically aligned causal latent modeling.} We introduce MAC-VAE, a motion–language–aligned causal VAE that learns temporally causal and semantically meaningful latent representations for text-to-motion generation.

\item \textbf{Frame-wise sampling with causal uncertainty.} We design a novel frame-wise sampling schedule that models causal uncertainty, allowing each frame to be predicted from partially denoised preceding frames for efficient, low-latency, and temporally consistent motion synthesis.

\item \textbf{Comprehensive empirical validation.} CMDM achieves state-of-the-art performance on HumanML3D and SnapMoGen, surpassing existing diffusion and autoregressive methods on text-to-motion generation and long-horizon motion generation.
\end{itemize}
\section{Related Works}
\label{sec:related}
\subsection{Motion-Language Alignment}
Recent advances in vision–language models~\cite{radford2021learning, oquab2023dinov2} have shown that large-scale training can robustly align text and visual semantics. This has led to a surge of interest in exploring motion–language alignment to enable practical control of motion using natural language. MotionCLIP~\cite{tevet2022motionclip} maps a single frame to CLIP space but fails to capture temporal dynamics. Subsequent methods, including TMR~\cite{petrovich2023tmr} and MotionPatches~\cite{yu2024exploring}, learn joint motion–text embeddings via contrastive or generative objectives, while PartTMR~\cite{yu2025remogpt} introduces body-part-level features for finer alignment. However, most motion–language models focus on retrieval tasks. Methods such as ReMoGPT~\cite{yu2025remogpt} and ReMoMask~\cite{li2025remomask} extend to text-to-motion generation but rely on retrieval-augmented generation rather than integrating motion–language alignment directly into the generation.

\subsection{Diffusion-based Motion Generation}
Text-conditioned motion generation has been explored through both non-diffusion and diffusion~\cite{ho2020denoising, ho2022classifier, dhariwal2021diffusion} paradigms. Early works used CNN- or RNN-based architectures~\cite{yan2019convolutional,zhao2020bayesian} and action-conditioned frameworks~\cite{guo2020action2motion,petrovich2021action} to synthesize motion from predefined semantics. More recently, diffusion-based methods~\cite{dhariwal2021diffusion,rombach2022high} have set new benchmarks for motion realism and diversity~\cite{zhang2022motiondiffuse,chen2023executing,mdm2022human}. MDM~\cite{mdm2022human} and MotionDiffuse~\cite{zhang2022motiondiffuse} operate directly in motion space, while MLD~\cite{chen2023executing}, MotionLCM~\cite{dai2024motionlcm}, EnergyMoGen~\cite{zhang2025energymogen} and SALAD~\cite{hong2025salad} perform diffusion in a latent space for greater stability and efficiency. However, these diffusion models rely on \emph{bidirectional} attention over entire sequences, breaking temporal causality and limiting real-time or streaming generation.

\subsection{Autoregressive Motion Generation}
Autoregressive (AR) modeling enforces temporal causality by predicting future frames from past context. Discrete-token methods such as T2M-GPT~\cite{zhang2023t2m} and MotionGPT~\cite{jiang2023motiongpt} treat motion as ``language,'' enabling powerful sequence modeling but suffering from exposure bias and cumulative errors. VQ-VAE-based approaches, including MoMask~\cite{guo2024momask}, MMM~\cite{pinyoanuntapong2024mmm}, and ParCo~\cite{zou2024parco}, quantize motion into discrete tokens and predict them autoregressively. Recent works explore causal paradigms for streaming generation: Dart~\cite{zhao2024dartcontrol} predicts short future segments from limited two historical frames, while MARDM~\cite{meng2024rethinking} and MotionStreamer~\cite{xiao2025motionstreamer} employ masked autoregressive transformers~\cite{li2024autoregressive} with diffusion heads. However, their reliance on teacher forcing~\cite{williams1989learning} and large diffusion heads causes instability in long-horizon inference and high computational cost, limiting real-time deployment.

Our work differs from existing methods in two key aspects: (1) we introduce a causal diffusion process within a motion–language–aligned latent space, preserving semantic consistency while enforcing temporal causality; and (2) we design a frame-wise sampling schedule that enables high-quality, streaming motion generation.
\section{Method}
\label{sec:method}

\begin{figure*}[t]
    \centering
    \includegraphics[width=1\linewidth]{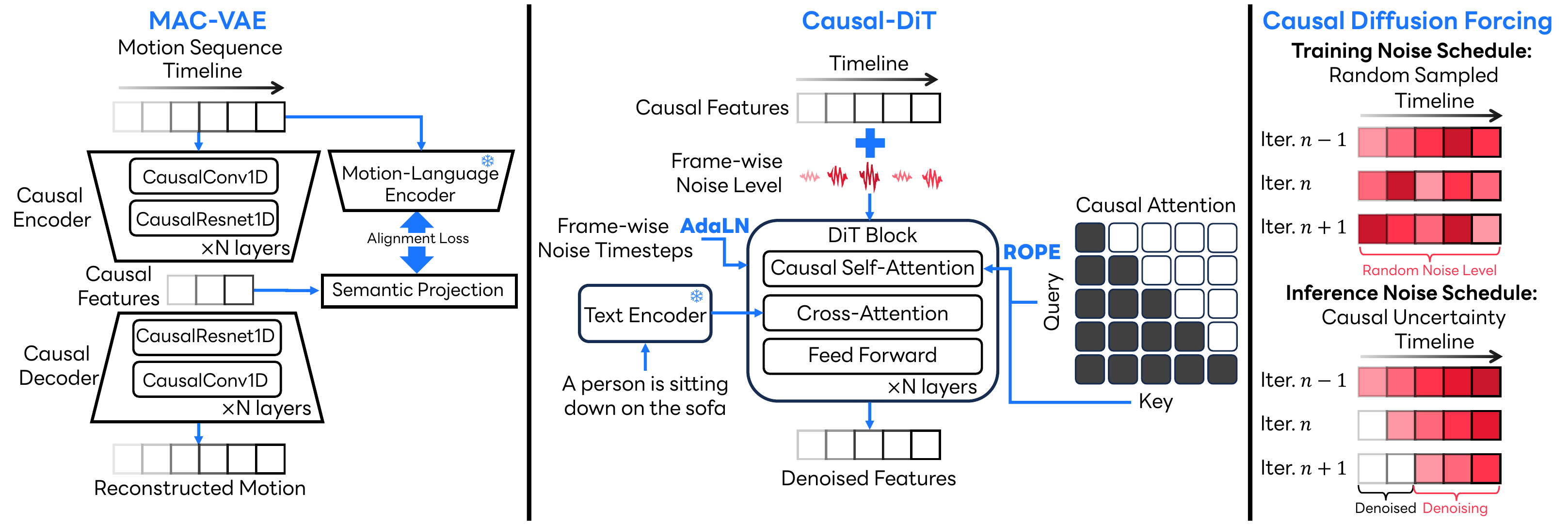}
    \vspace{-15pt}
    \caption{Overview of the proposed CMDM framework. CMDM consists of three key components: (a) MAC-VAE, which encodes motion sequences into motion–language–aligned and temporally causal latent features using a causal encoder–decoder structure supervised by motion-language model alignment; (b) Causal-DiT, which performs diffusion denoising with causal self-attention and cross-attention to text embeddings, ensuring temporally ordered and semantically consistent frame refinement; and (c) Causal Diffusion Forcing, which applies independent frame-level noise during training and a causal uncertainty schedule during inference, where the {\color{red}{redness intensity}} represents the noise level. This design enables CMDM to achieve temporally consistent, semantically aligned, and efficient text-to-motion generation suitable for streaming and long-horizon synthesis.}
    \vspace{-10pt}
    \label{fig:framework}
\end{figure*}

Our proposed framework, Causal Motion Diffusion Models (CMDM), enables temporally ordered, text-conditioned motion generation by integrating causal latent encoding, causal diffusion forcing, and efficient frame-wise sampling.
As illustrated in~\Fref{fig:framework}, CMDM consists of three core components:
(1) a Motion-Language-Aligned Causal VAE (MAC-VAE) that encodes motion sequences into semantically aligned and temporally causal latent spaces,
(2) a Causal Diffusion Transformer (Causal-DiT) that performs frame-wise diffusion with causal self-attention to ensure autoregressive temporal dependencies, and
(3) a Frame-Wise Sampling Scheduler (FSS) that models causal uncertainty by assigning higher noise to future frames and lower noise to past frames, allowing each new frame to be predicted from partially denoised preceding frames.

\subsection{Motion-Language-Aligned Causal VAE}

To obtain a temporally structured and semantically consistent latent representation, CMDM employs a \emph{causal variational autoencoder} aligned with a motion-language foundation model.
Given a motion sequence $\mathbf{x}_{1:T} \in \mathbb{R}^{T \times D}$, where $T$ is the number of frames and $D$ is the dimension of the joint representation, the proposed MAC-VAE encoder $E_\phi$ and decoder $D_\psi$ operate causally such that each frame depends only on its past:
\begin{equation}
\mathbf{z}_t = E_\phi(\mathbf{x}_{\le t}), \quad
\hat{\mathbf{x}}_t = D_\psi(\mathbf{z}_{\le t}),
\end{equation}
where $\mathbf{z}_t \in \mathbb{R}^{d_z}$ denotes the latent representation at timestep $t$.
For a motion sequence with $T$ frames, we encode it into $T/4$ temporal steps in the latent space, effectively achieving a $4\times$ temporal downsampling ratio.
This compression balances representation compactness and temporal resolution, reducing redundancy while preserving the underlying motion dynamics.

The encoder and decoder are adapted from~\cite{xiao2025motionstreamer} and are composed of 1D causal convolution and 1D causal ResNet blocks, ensuring temporal causality during both encoding and reconstruction.
In this design, each frame only depends on preceding frames, while future frames are excluded from computation, explicitly modeling temporal causality in the latent space.
During inference, reconstructed motions can be decoded sequentially in real time, enabling streaming generation without requiring access to future frames.

To enhance semantic alignment, motion features are projected through a pretrained motion--language encoder (Part-TMR~\cite{yu2025remogpt}), which provides part-level semantic supervision.
The MAC-VAE objective combines three terms: the standard VAE reconstruction loss, the Kullback--Leibler divergence, and a newly introduced motion--language alignment loss:
\begin{equation}
\mathcal{L}_{\text{MAC-VAE}} =
\mathcal{L}_{\text{rec}}
+ \beta D_{\text{KL}}\big(q_\phi(\mathbf{z}|\mathbf{x}) \,\|\, p(\mathbf{z})\big)
+ \lambda \mathcal{L}_{\text{align}}.
\end{equation}

\paragraph{Motion Alignment Loss.}
To enforce fine-grained semantic alignment between motion and text, we introduce a motion alignment loss $\mathcal{L}_{\text{align}}$ that measures both point-to-point feature similarity and relative structural consistency between motion embeddings $\mathbf{z}$ and motion-language features $\mathbf{f}$ extracted from the motion encoder of the pretrained Part-TMR model~\cite{yu2025remogpt}.
Specifically, we employ two complementary objectives, following the design in VAVAE~\cite{yao2025reconstruction}: (1) a \textit{marginal cosine similarity loss} that minimizes local feature gaps, and (2) a \textit{marginal distance matrix similarity loss} that preserves the relational geometry of feature spaces, as defined below:
\begin{equation}
\mathcal{L}_{\text{align}} =
\mathcal{L}_{\text{mcos}} +
\mathcal{L}_{\text{mdms}}.
\end{equation}

Given aligned feature maps $\mathbf{Z}$ and $\mathbf{F}$ from the latent space and the motion--language encoder, respectively, we project $\mathbf{Z}$ to match the feature dimensionality of $\mathbf{F}$ via a linear transformation:
\begin{equation}
\mathbf{Z}' = W \mathbf{Z},
\end{equation}
where $W \in \mathbb{R}^{d_f \times d_z}$ is a learnable projection matrix.
The marginal cosine similarity loss $\mathcal{L}_{\text{mcos}}$ minimizes the similarity gap between corresponding features $\mathbf{z}'_{ij}$ and $\mathbf{f}_{ij}$:
\begin{equation}
\mathcal{L}_{\text{mcos}} = 
\frac{1}{N}
\sum_{i,j}
\text{ReLU}\!\left( 1 - m_1 - 
\frac{\mathbf{z}'_{ij} \cdot \mathbf{f}_{ij}}{\|\mathbf{z}'_{ij}\| \, \|\mathbf{f}_{ij}\|} \right),
\end{equation}
where $N$ is the total number of temporal feature elements, and $m_1$ is a similarity margin that encourages stronger alignment for less similar pairs.
This loss focuses learning on semantically misaligned regions, improving feature-level consistency.

Complementary to $\mathcal{L}_{\text{mcos}}$, the marginal distance matrix similarity loss $\mathcal{L}_{\text{mdms}}$ enforces the alignment of internal structural relationships between motion and text embeddings by matching their pairwise distance matrices.
Formally, we compute:
\begin{equation}
\mathcal{L}_{\text{mdms}} =
\frac{1}{N^2}
\sum_{i,j}
\text{ReLU}\!\left(
\left|
\frac{\mathbf{z}_i \cdot \mathbf{z}_j}{\|\mathbf{z}_i\|\|\mathbf{z}_j\|} -
\frac{\mathbf{f}_i \cdot \mathbf{f}_j}{\|\mathbf{f}_i\|\|\mathbf{f}_j\|}
\right|
- m_2 \right),
\end{equation}
where $m_2$ is a distance margin that relaxes the alignment constraint for similar pairs.
This objective promotes structural consistency between latent and text spaces, ensuring that the relative geometry of motion embeddings matches that of the aligned foundation features.

\subsection{Causal Diffusion Forcing}

We extend Diffusion Forcing~\cite{chen2024diffusion}, originally proposed for next-token prediction, to the motion domain to model autoregressive temporal dynamics in the latent space.
Unlike conventional diffusion models that jointly denoise all frames, our CMDM introduces \emph{frame-level noise} with independent diffusion timesteps for each motion frame in the latent sequence, enforcing causal dependencies between past and future frames.

In standard full-sequence diffusion, the same level of noise $k \in [0, K]$, where $K$ is the total number of diffusion steps, is applied to the entire sequence as:
\begin{equation}
\tilde{\mathbf{z}}^k =
\sqrt{\bar{\alpha}_{k}}\, \mathbf{z}^k +
\sqrt{1 - \bar{\alpha}_{k}}\, \boldsymbol{\epsilon}^k,
\quad
\boldsymbol{\epsilon}^k \sim \mathcal{N}(0, I),
\end{equation}
and the denoising model $\epsilon_\theta$ is trained to recover the original sequence simultaneously:
\begin{equation}
\mathcal{L} =
\mathbb{E}_{k, \boldsymbol{\epsilon}^k}
\Big[
\| \boldsymbol{\epsilon}^k -
\epsilon_\theta(\tilde{\mathbf{z}}^k, k, \mathbf{c}) \|_2^2
\Big],
\end{equation}
where $\mathbf{c}$ denotes the text embedding.

In \emph{causal diffusion forcing}, each frame $t$ is assigned an independent noise level $k_t \in [0, K]$, and the noisy latent representation is defined as:
\begin{equation}
\tilde{\mathbf{z}}_t^{k_t} =
\sqrt{\bar{\alpha}_{k_t}}\, \mathbf{z}_t^{k_t} +
\sqrt{1 - \bar{\alpha}_{k_t}}\, \boldsymbol{\epsilon}_t^{k_t},
\quad
\boldsymbol{\epsilon}_t^{k_t} \sim \mathcal{N}(0, I).
\end{equation}
The diffusion transformer $\epsilon_\theta$ predicts the noise residual using \emph{causal self-attention}, which restricts each frame to attend only to its past representations:
\begin{equation}
\mathcal{L}_{\text{DF}} =
\mathbb{E}_{k_t, \boldsymbol{\epsilon}_t^{k_t}}
\Big[
\| \boldsymbol{\epsilon}_t^{k_t} -
\epsilon_\theta(\tilde{\mathbf{z}}_{\le t}, k_t, \mathbf{c}) \|_2^2
\Big].
\end{equation}

This causal training setup ensures that the denoising process evolves forward in time, with each prediction depending solely on the available history, effectively bridging diffusion and autoregression.

By replacing the globally synchronized noise schedule with frame-specific perturbations, CMDM achieves several advantages.
First, the model learns to operate under diverse noise conditions at each frame, which improves temporal robustness and generalization to variable-length sequences.
Second, the causal attention mask explicitly enforces temporal order within the transformer backbone, preventing information leakage from future frames and enabling real-time or streaming generation.
Finally, the per-frame stochasticity of diffusion forcing acts as a natural regularizer, encouraging smooth temporal transitions while preserving motion diversity.

\subsection{Causal Diffusion Transformer}

To model the temporal dependencies of diffusion forcing, CMDM employs a causal diffusion transformer (Causal-DiT) that performs diffusion-based denoising under strict causal constraints. Unlike conventional transformers with bidirectional attention, Causal-DiT uses causal masking so each frame accesses only its past and current context, ensuring sequential generation consistent with autoregressive reasoning while maintaining diffusion fidelity.

Each transformer block integrates three key mechanisms: (1) causal self-attention, which employs a lower-triangular attention mask to prevent each frame from attending to future frames. This constraint preserves the causal order required for autoregressive modeling and ensures that the model predicts future motion dynamics based solely on previously observed information. (2) cross-attention, which establishes the link between motion and language by conditioning frame-level motion latents on word-level text embeddings extracted from DistilBERT~\cite{sanh2019distilbert}. Through this mechanism, semantic cues from natural language guide the temporal evolution of motion features, allowing the model to synthesize actions that remain coherent with the textual descriptions across the sequence. (3) adaptive layer normalization (AdaLN)~\cite{peebles2023scalable} combined with rotary positional encoding (ROPE)~\cite{su2024roformer}, where AdaLN embeds frame-wise diffusion timestep information, ensuring that temporal noise levels $k_t$ are seamlessly integrated into the denoising process, while ROPE stabilizes long-horizon denoising with relative positional encoding.

During training, each frame $t$ is diffused with an independent noise level $k_t$, and the model learns to denoise them as:
\begin{equation}
\epsilon_\theta(\tilde{\mathbf{z}}_{\le t}, k_t, \mathbf{c}) =
\text{CausalDiT}(\tilde{\mathbf{z}}_{\le t}, k_t, \mathbf{c}),
\end{equation}
where $\tilde{\mathbf{z}}_{\le t}$ denotes the partially noised causal latent sequence and $\mathbf{c}$ is the text embedding.

\subsection{Inference and Streaming Generation}

During inference, CMDM generates motion autoregressively by progressively denoising each frame in a causal manner. 
Given text condition $\mathbf{c}$ and an initial noise sequence $\{\tilde{\mathbf{z}}_t^{K}\}_{t=1}^{T} \sim \mathcal{N}(0, I)$, the model predicts each frame conditioned on previously denoised latents:
\begin{equation}
\tilde{\mathbf{z}}_t^{k_t-1} = G_\theta(\{\tilde{\mathbf{z}}^0_{<t}, \tilde{\mathbf{z}}_t^{k_t}\}, k_t, \mathbf{c}),
\quad
\hat{\mathbf{x}}_t = D_\psi(\tilde{\mathbf{z}}^0_{\le t}),
\end{equation}
where $G_\theta$ denotes the causal diffusion generator.
This formulation ensures strictly causal synthesis and, with key-value caching for autoregressive rollout, enables real-time generation. However, this scheme is prone to accumulating single-step errors, as it treats the predicted $\tilde{\mathbf{z}}_t^0$ as a ground truth observation, a practice more broadly referred to as exposure bias~\cite{schmidt2019generalization, ning2023elucidating}.

\paragraph{Frame-wise Sampling Schedule (FSS).}
To accelerate inference and mitigate exposure bias, CMDM adopts a frame-wise sampling schedule with \emph{causal uncertainty}, assigning lower noise to past frames and higher noise to future ones.
At each step, the model refines the next frame using partially denoised histories. For example, a causal uncertainty schedule with uncertainty scale $L$ can be defined as:

\begin{equation}
K_{m,t} =
\begin{bmatrix}
K & K & K \\
K{-}L & K & K \\
K{-}2L & K{-}L & K \\
\vdots & \ddots & K{-}L  \\
0 & \cdots & \vdots \\
0 & 0 & 0
\end{bmatrix},
\end{equation}
where $K_{m,t}$ indicates the noise applied to frame $t$ at iteration $m$ and the uncertainty scale $L$ indicates that the denoising of the next frame begins at step $K{-}L$ of the current frame. Intermediate steps between $K$, $K{-}L$, $K{-}2L$, and so on are omitted for clarity. Each partially denoised frame $\tilde{\mathbf{z}}_t$ is reused as context for subsequent predictions, enabling continuous, low-latency generation with high temporal coherence and smooth transitions, while greatly reducing inference cost compared to full autoregressive diffusion.

\section{Experiments}
\label{sec:experiments}

\begin{table*}[t]
    \centering
    \resizebox{0.95\linewidth}{!}{
    \begin{tabular}{l| c | c c c| c |c| c| c}
    \toprule
      \multirow{1}{*}{Methods} & \multirow{2}{*}{Framework}  & \multicolumn{3}{c|}{R-Precision$\uparrow$} & \multirow{2}{*}{FID$\downarrow$} & \multirow{2}{*}{MM-Dist$\downarrow$} & \multirow{2}{*}{MModality$\uparrow$} & \multirow{2}{*}{CLIP-score$\uparrow$}\\
    \cline{3-5}
     ~ & ~& Top 1 & Top 2 & Top 3 & & & & \\
    \midrule\midrule 
    GT & $-$
    & $0.511^{\pm.003}$ & $0.703^{\pm.003}$ & $0.797^{\pm.002}$ 
    & $0.002^{\pm.000}$ & $2.974^{\pm.008}$ & $-$ & $0.639^{\pm.001}$  \\
    \midrule
    T2M-GPT~\cite{zhang2023t2m} & \multirow{3}{*}{VQ} 
    & $0.492^{\pm.003}$ &$0.679^{\pm.002}$  & $0.775^{\pm.002}$ 
    & $0.141^{\pm.005}$ & $3.121^{\pm.009}$& $1.831^{\pm.048}$ & $0.607^{\pm.005}$ \\
    
    MMM~\cite{pinyoanuntapong2024mmm} &  
    & $0.515^{\pm.002}$& $0.708^{\pm.002}$& $0.804^{\pm.002}$ 
    & $0.089^{\pm.005}$& $2.926^{\pm.007}$& $1.226^{\pm.035}$ & $0.635^{\pm.003}$ \\ 
    
    MoMask~\cite{guo2024momask} & 
    & $0.521^{\pm.002}$ & $0.713^{\pm.002}$  &$0.807^{\pm.002}$ 
    & $\mathbf{0.045^{\pm.002}}$ &$2.958^{\pm.008}$& $1.241^{\pm.040}$ & ${0.637^{\pm.003}}$\\
    \midrule
    
    MDM-50Steps~\cite{mdm2022human} & \multirow{6}{*}{Diffusion} 
    & $0.455^{\pm.006}$ & $0.645^{\pm.007}$& $0.749^{\pm.002}$ 
    & $0.489^{\pm.025}$ & $3.330^{\pm.025}$& $\mathbf{2.290^{\pm.070}}$ & $0.481^{\pm.001}$\\
    
    MLD V2~\cite{chen2023executing} &  
    & $0.542^{\pm.002}$ &$0.735^{\pm.002}$&$0.827^{\pm.002}$ 
    & $0.078^{\pm.004}$& $2.808^{\pm.006}$& $1.676^{\pm.060}$ & $0.645^{\pm.003}$\\
    
    MotionLCM V2~\cite{dai2024motionlcm} &  
    & $0.548^{\pm.002}$ &$0.743^{\pm.002}$&$0.835^{\pm.002}$ 
    & $0.092^{\pm.003}$& $2.760^{\pm.008}$& $1.800^{\pm.047}$ & $0.646^{\pm.003}$\\
    
    StableMoFusion~\cite{huang2024stablemofusion} &
    & $0.553^{\pm.003}$ &$0.748^{\pm.002}$&$0.841^{\pm.002}$ 
    & $0.098^{\pm.003}$& $2.715^{\pm.006}$ & $1.774^{\pm.051}$ & $0.651^{\pm.001}$\\

    EnergyMoGen~\cite{zhang2025energymogen} &
    & $0.523^{\pm.003}$ & $0.715^{\pm.002}$ & $0.815^{\pm.002}$ & $0.188^{\pm.006}$ & $2.915^{\pm.007}$ & $\underline{2.205^{\pm.041}}$ & $-$ \\
    
    SALAD~\cite{hong2025salad} &
    & $\underline{0.581^{\pm.003}}$ & $\underline{0.769^{\pm.003}}$ & $\underline{0.857^{\pm.002}}$ & ${0.076^{\pm.002}}$ & $2.649^{\pm.009}$ & $1.751^{\pm.062}$ & $0.671^{\pm.001}$ \\
    \midrule
    
    MARDM$^\dagger$~\cite{meng2024rethinking} & Autoregressive 
    & $0.517^{\pm.003}$ &$0.708^{\pm.003}$&$0.805^{\pm.003}$ 
    & $0.116^{\pm.006}$& $2.968^{\pm.010}$ & ${1.923^{\pm.105}}$ & $0.642^{\pm.001}$\\
    
    MotionStreamer$^\dagger$~\cite{xiao2025motionstreamer} & Transformer 
    & $0.496^{\pm.003}$ &$0.695^{\pm.002}$&$0.793^{\pm.002}$ 
    & $0.201    ^{\pm.005}$& $3.041^{\pm.009}$ & ${1.463^{\pm.075}}$ & $0.610^{\pm.002}$ \\
    \midrule
    
    \textbf{CMDM w/ AR} &  Autoregressive
    & ${0.576^{\pm.003}}$ & ${0.768^{\pm.004}}$& ${0.853^{\pm.002}}$ 
    & ${0.079^{\pm.003}}$ & $\underline{2.647^{\pm.011}}$& ${1.951^{\pm.086}}$ & $\underline{0.675^{\pm.001}}$ \\
    
    \textbf{CMDM w/ FSS} & Diffusion
    & $\mathbf{0.588^{\pm.004}}$ & $\mathbf{0.778^{\pm.002}}$& $\mathbf{0.860^{\pm.003}}$ 
    & $\underline{0.068^{\pm.003}}$ & $\mathbf{2.620^{\pm.010}}$& ${1.785^{\pm.074}}$ & $\mathbf{0.685^{\pm.001}}$  \\
    \bottomrule
    \end{tabular}}
    \vspace{-7pt}
    \caption{Results of text-to-motion generation on HumanML3D. The average is reported over 10 runs with 95\% confidence intervals. Methods marked with $\dagger$ were originally implemented with different motion representations and have been re-trained using our codebase to ensure a fair comparison. \textbf{Bold} indicates the best result, while \underline{underline} denotes the second-best result.}
    \vspace{-13pt}
    \label{tab:result}
\end{table*}

\subsection{Experimental Setup}
\paragraph{Datasets.}
To evaluate CMDM, we conduct experiments on two benchmarks: HumanML3D~\cite{Guo_2022_CVPR_humanml3d} and SnapMoGen~\cite{guo2025snapmogen}.
HumanML3D contains 14,616 motion clips from AMASS~\cite{AMASS_ICCV2019} paired with 44,970 short textual descriptions of common actions (\eg, walking, jumping, sitting).
SnapMoGen includes 20,450 motion capture clips with 122K expressive captions (average 48 words) covering about 43.7 hours of data.
Unlike HumanML3D, SnapMoGen features temporally continuous, long-horizon activities (\eg, sports and performances), allowing evaluation of smooth, consistent motion generation.
For fair comparison, we follow the standard 3D motion representation of each dataset: 263 dimensions for HumanML3D and 296 for SnapMoGen, including joint velocities, positions, and rotations.

\paragraph{Evaluation Metrics.}
Following prior work~\cite{Guo_2022_CVPR_humanml3d, jiang2023motiongpt, guo2025snapmogen}, we evaluate CMDM using several standard metrics.
(1) Motion quality is measured by Fréchet Inception Distance (FID), which assesses the realism of generated motions relative to ground truth.
(2) Multi-modality quantifies the diversity of motions generated from identical text prompts.
(3) Text–motion alignment is evaluated by R-Precision (R@1, R@2, R@3) and Multi-Modal Distance (MM Dist) using a pretrained text–motion retrieval model.
We also report the CLIP-Score~\cite{meng2024rethinking}, which measures the cosine similarity between generated motion features and their corresponding captions in the CLIP embedding space.

\paragraph{Implementation Details.}
MAC-VAE comprises seven causal convolution layers and two causal ResNet blocks with left padding in both the encoder and decoder. The latent feature dimension is set to 64. We modify and retrain Part-TMR~\cite{yu2025remogpt} to extract frame-level semantic motion–language features for supervising MAC-VAE. The loss weight $\lambda$ for semantic alignments is automatically adjusted according to the gradient norm at the final layer of the encoder to maintain balance with other losses. The Causal-DiT is implemented as a lightweight Transformer~\cite{vaswani2017attention} with eight layers, four attention heads, and a hidden dimension of 512. Flow Matching~\cite{lipman2022flow, ma2024sit, albergo2023stochastic, albergo2022building} is adopted as the ODE sampler for causal diffusion forcing. For FSS, we set 50 denoising step with $K=50$ for each frame and start to denoise the next frame at $K-2$ (\ie, $L=2$) during inference.

\begin{table*}[t]
\centering
\resizebox{0.92\linewidth}{!}{
\begin{tabular}{l|c| ccc|c|c|c}
\toprule
\multirow{2}{*}{Methods} & \multirow{2}{*}{Framework} &  \multicolumn{3}{c|}{R-Precision$\uparrow$} & \multirow{2}{*}{FID$\downarrow$} & \multirow{2}{*}{MModality$\uparrow$} & \multirow{2}{*}{CLIP-score$\uparrow$} \\
\cline{3-5}
~ & ~ & Top 1 & Top 2 & Top 3 & ~ & ~ & ~ \\
\midrule\midrule 
GT & $-$ & $0.940^{\pm.001}$ & $0.976^{\pm.001}$ & $0.985^{\pm.001}$ 
& $0.001^{\pm.000}$ & $ - $ & $0.837^{\pm.000}$ \\ \midrule
T2M-GPT~\cite{zhang2023t2m} & \multirow{3}{*}{VQ} 
& $0.618^{\pm.002}$ & $0.773^{\pm.002}$ & $0.812^{\pm.002}$ 
& $32.629^{\pm.087}$ & $9.172^{\pm.181}$ & $0.573^{\pm.001}$ \\
MoMask~\cite{guo2024momask} & 
& $0.777^{\pm.002}$ & $0.888^{\pm.002}$ & $0.927^{\pm.002}$ 
& $17.404^{\pm.051}$ & $8.183^{\pm.184}$ & $0.664^{\pm.001}$ \\
MoMask$^{++}$~\cite{guo2025snapmogen} & 
& $0.802^{\pm.001}$ & ${0.905^{\pm.002}}$ & ${0.938^{\pm.001}}$ 
& ${15.061^{\pm.065}}$ & $7.259^{\pm.180}$ & ${0.685^{\pm.001}}$ \\
\midrule
MDM~\cite{mdm2022human} & \multirow{2}{*}{Diffusion} 
& $0.503^{\pm.002}$ & $0.653^{\pm.002}$ & $0.727^{\pm.002}$ 
& $57.783^{\pm.092}$ & $\mathbf{13.412^{\pm.231}}$ & $0.481^{\pm.001}$ \\
StableMoFusion~\cite{huang2024stablemofusion} & 
& $0.679^{\pm.002}$ & $0.823^{\pm.002}$ & $0.888^{\pm.002}$ 
& $27.801^{\pm.063}$ & $9.064^{\pm.138}$ & $0.605^{\pm.001}$ \\
\midrule
MARDM$^\dagger$~\cite{meng2024rethinking} & Autoregressive
& $0.648^{\pm.002}$ & $0.801^{\pm.002}$ & $0.856^{\pm.002}$ 
& $26.348^{\pm.208}$ & $\underline{9.883^{\pm.147}}$ & $0.601^{\pm.001}$ \\
MotionStreamer$^\dagger$~\cite{xiao2025motionstreamer} & Transformer 
& $0.631^{\pm.002}$ & $0.791^{\pm.002}$ & $0.836^{\pm.002}$ 
& $30.023^{\pm.131}$ & ${7.543^{\pm.195}}$ & $0.580^{\pm.001}$ \\
\midrule
\textbf{CMDM w/ AR} & Autoregressive & $\underline{0.824^{\pm.002}}$ & $\underline{0.918^{\pm.004}}$ & $\underline{0.951^{\pm.002}}$ 
& $\underline{15.008^{\pm.074}}$ & $9.735^{\pm.186}$ & $\underline{0.699^{\pm.001}}$ \\
\textbf{CMDM w/ FSS} & Diffusion  & $\mathbf{0.831^{\pm.004}}$ & $\mathbf{0.926^{\pm.003}}$ & $\mathbf{0.958^{\pm.002}}$ 
& $\mathbf{14.451^{\pm.074}}$ & $9.521^{\pm.196}$ & $\mathbf{0.702^{\pm.001}}$ \\
\bottomrule
\end{tabular}}
\vspace{-7pt}
\caption{Results of text-to-motion generation on SnapMoGen. The average is reported over 10 runs with 95\% confidence intervals.
}
\vspace{-7pt}
\label{tab:sanp_results}
\end{table*}

\begin{table*}[t]
\centering
\resizebox{0.9\linewidth}{!}{
\begin{tabular}{l|l|cccc|cccc}
\toprule
& \multirow{2}{*}{Methods}  & \multicolumn{4}{c|}{Subsequence} & \multicolumn{4}{c}{Transition} \\ 
\cline{3-10}
& ~ & R-Top3$\uparrow$ & FID$\downarrow$ & Div$\rightarrow$ & MM-Dist$\downarrow$ & FID$\downarrow$ & Div$\rightarrow$ & PJ$\rightarrow$ & AUJ$\downarrow$ \\ 
\midrule \midrule 
\multirow{4}{*}{\rotatebox{90}{\textbf{HML3D}}} & GT & $0.796^{\pm0.004}$ & $0.00^{\pm0.00}$ & $9.34^{\pm0.08}$ & $2.97^{\pm0.01}$ & $0.00^{\pm0.00}$ & $9.54^{\pm0.15}$ & $0.04^{\pm0.00}$ & $0.07^{\pm0.00}$ \\ \cmidrule{2-10}
& FlowMDM~\cite{barquero2024seamless} & ${0.685^{\pm0.004}}$ & ${0.29^{\pm0.01}}$ & ${9.58^{\pm0.12}}$ & ${3.61^{\pm0.01}}$ & $\mathbf{1.38^{\pm0.05}}$ & $\mathbf{8.79^{\pm0.09}}$ & $\underline{0.06^{\pm0.00}}$ & $\underline{0.51^{\pm0.01}}$ \\
& MARDM~\cite{meng2024rethinking}  & $\underline{{0.741^{\pm0.004}}}$ & $\underline{{0.24^{\pm0.01}}}$ & $\underline{{9.29^{\pm0.12}}}$ & $\underline{3.38^{\pm0.01}}$ & ${2.72^{\pm0.09}}$ & $8.28^{\pm0.11}$ & ${0.07^{\pm0.00}}$ & ${0.57^{\pm0.01}}$ \\
\rowcolor[gray]{0.90}
\cellcolor{white} & \textbf{CMDM} & $\mathbf{0.782^{\pm0.003}}$ & $\mathbf{0.12^{\pm0.04}}$ & $\mathbf{9.44^{\pm0.13}}$ & $\mathbf{3.04^{\pm0.01}}$ & $\underline{1.66^{\pm0.06}}$ & $\underline{8.72^{\pm0.10}}$ & $\mathbf{0.04^{\pm0.00}}$ & $\mathbf{0.42^{\pm0.01}}$ \\
\toprule
 \multirow{4}{*}{\rotatebox{90}{\textbf{SnapMG}}} & GT & $0.997^{\pm0.006}$ & $0.00^{\pm0.00}$ & $19.74^{\pm0.05}$ & $14.83^{\pm0.01}$ & $0.00^{\pm0.00}$ & $19.21^{\pm0.08}$ & $1.11^{\pm0.01}$ & $45.16^{\pm0.33}$ \\ \cmidrule{2-10}
& FlowMDM~\cite{barquero2024seamless} & $0.485^{\pm0.009}$ & $69.10^{\pm0.70}$ & $19.54^{\pm0.07}$ & $31.85^{\pm0.08}$ & $62.91^{\pm0.63}$ & $\underline{19.04^{\pm0.08}}$ & $\mathbf{0.91^{\pm0.03}}$ & $\mathbf{23.01^{\pm1.68}}$ \\
& MARDM~\cite{meng2024rethinking}  & $\underline{0.644^{\pm0.007}}$ & $\underline{40.80^{\pm0.17}}$ & $\underline{19.68^{\pm0.06}}$ & $\underline{28.93^{\pm0.10}}$ & $\underline{54.10^{\pm0.46}}$ & $\mathbf{19.13^{\pm0.07}}$ & $8.12^{\pm0.23}$ & $130.33^{\pm1.08}$ \\
\rowcolor[gray]{0.90}
\cellcolor{white} & \textbf{CMDM} & $\mathbf{0.852^{\pm0.003}}$ & $\mathbf{32.49^{\pm0.16}}$ & $\mathbf{19.80^{\pm0.07}}$ & $\mathbf{24.94^{\pm0.05}}$ & $\mathbf{38.73^{\pm0.65}}$ & $19.50^{\pm0.08}$ & $\underline{2.54^{\pm0.06}}$ & $\underline{70.35^{\pm0.68}}$ \\
\bottomrule
\end{tabular}}
\vspace{-7pt}
\caption{Results of long-horizon motion generation on HumanML3D and SnapMoGen. The motion quality of each subsequence and the smoothness of each transition are evaluated.  }
\vspace{-13pt}
\label{tab:long_comparison}
\end{table*}

\begin{figure*}[t]
    \centering
    \includegraphics[width=1\linewidth]{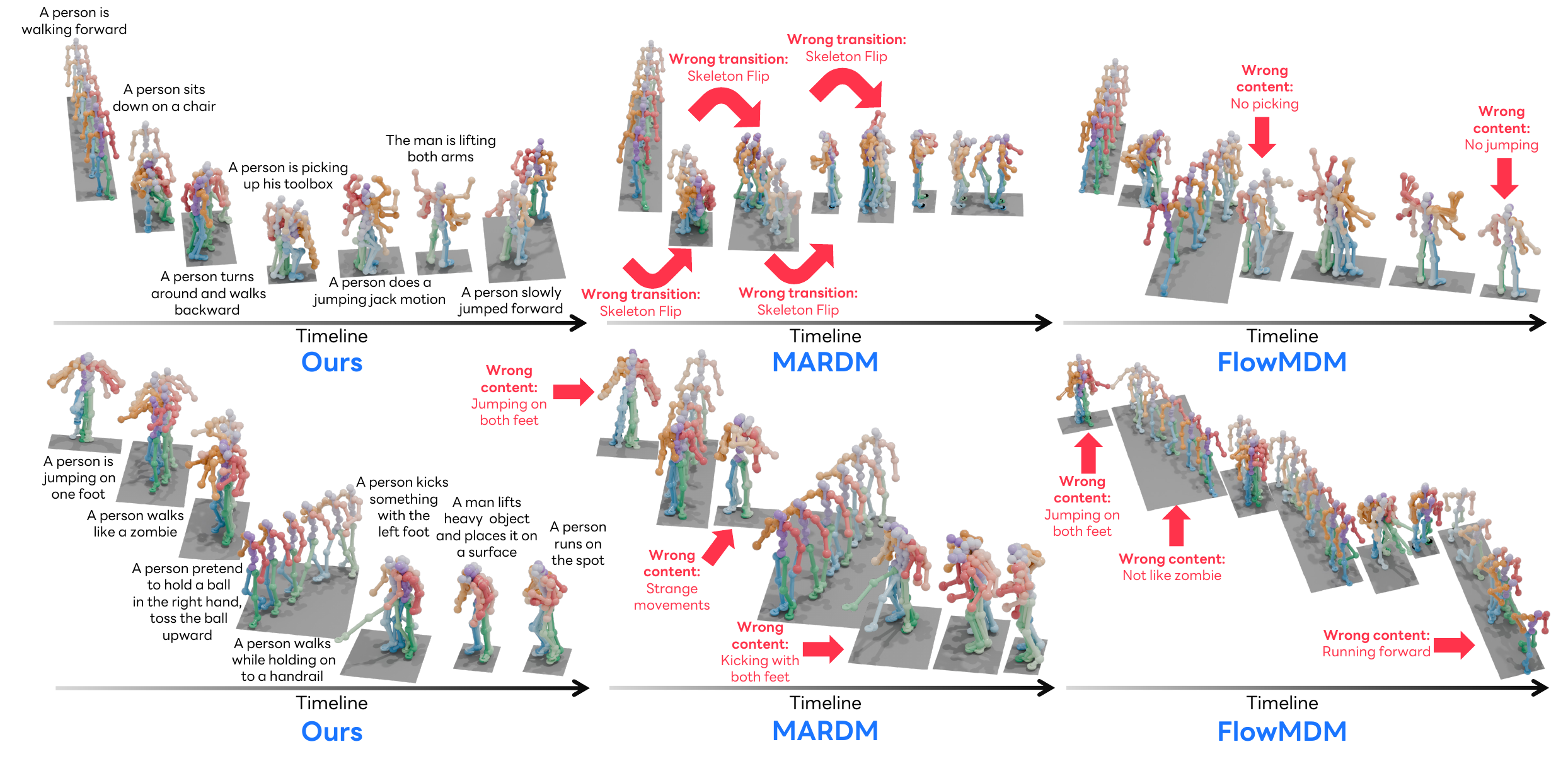}
    \vspace{-23pt}
    \caption{Qualitative results of long-horizon motion generation. Comparison between our CMDM and previous methods. The generated motion is continuous and seamless; for visualization purposes, we split each long sequence into shorter segments corresponding to their captions. Please refer to the videos in the supplementary materials for the complete motion sequences.}
    \vspace{-10pt}
    \label{fig:qualitative_long}
\end{figure*}

\subsection{Quantitative Results}
\paragraph{Results on HumanML3D.}
We compare CMDM with state-of-the-art models across three paradigms: (1) \emph{VQ-based}~\cite{zhang2023t2m, pinyoanuntapong2024mmm, guo2024momask}; (2) \emph{Diffusion-based} ~\cite{mdm2022human, chen2023executing, dai2024motionlcm, huang2024stablemofusion, zhang2025energymogen, hong2025salad}; and (3) \emph{Autoregressive-based}~\cite{meng2024rethinking, xiao2025motionstreamer}.
As shown in~\Tref{tab:result}, CMDM consistently achieves superior or comparable results across all metrics. CMDM with frame-wise sampling (CMDM w/ FSS) attains the best overall performance, achieving an R-Precision of 0.588/0.778/0.860 (Top-1/2/3), the second lowest FID (0.068), and the highest CLIP-Score (0.685). These results indicate high motion fidelity and strong text–motion alignment.
Compared to standard autoregressive sampling (CMDM w/ AR), FSS further improves temporal stability and smoothness while reducing inference latency. This improvement arises from the causal uncertainty mechanism, where each subsequent frame is generated from partially denoised preceding frames. This design allows the model to reduce the accumulated error of autoregressive and adaptively refine local temporal transitions while maintaining global coherence, leading to smoother motion dynamics and improved semantic alignment for stable generation.

\paragraph{Results on SnapMoGen.}
We further evaluate the proposed CMDM on the motion clips of SnapMoGen, which contains expressive motion sequences paired with rich textual descriptions. As shown in~\Tref{tab:sanp_results}, CMDM achieves state-of-the-art performance across all evaluation metrics, demonstrating its strong generalization to complex motions. CMDM with the frame-wise sampling schedule (CMDM w/ FSS) achieves the best overall results, surpassing all previous VQ-, diffusion-, and autoregressive-based methods. It also achieves the lowest FID score and a high CLIP-Score, indicating superior motion realism and semantic alignment.

\paragraph{Long-Horizon Motion Generation}
To evaluate CMDM on long-horizon motion generation, we compare it with the motion composition method FlowMDM~\cite{barquero2024seamless} and the autoregressive model MARDM~\cite{meng2024rethinking}.
Following the protocol of FlowMDM, we synthesize 64 long-horizon sequences on HumanML3D by composing 32 caption–duration pairs per sequence, evaluating 32 subsequences and 31 transitions for local quality and temporal continuity. The ground-truth metrics are computed using randomly sampled motion clips from HumanML3D.
For SnapMoGen, which provides ground-truth long sequences, we select 128 samples with over five continuous motions and use the same captions for generation. We further employ Peak Jerk (PJ) and Area Under the Jerk (AUJ)~\cite{barquero2024seamless} to measure transition smoothness.
Owing to differences in skeleton scale, the magnitude of the metrics differs from those reported on HumanML3D.
The results are shown in~\Tref{tab:long_comparison}. Although FlowMDM reports lower PJ and AUJ values on SnapMoGen, this is primarily because its generated motions often remain static or frozen as can be seen from the ensuing qualitative analysis in~\Fref{fig:qualitative_long}. In contrast, CMDM produces temporally consistent, smoothly transitioning, and realistic long-horizon motions at real-time speed, demonstrating its effectiveness for streaming and continuous text-to-motion generation.

\subsection{Qualitative Results}
\Fref{fig:qualitative_long} shows qualitative comparisons with FlowMDM~\cite{barquero2024seamless} and MARDM~\cite{meng2024rethinking} on long-horizon motion generation. Given a sequence of captions, CMDM generates continuous and seamless motions with accurate semantics and smooth transitions across segments. In contrast, FlowMDM and MARDM often produce incorrect actions, \eg, ``no jumping'' or unnatural transitions, \eg, skeleton flips. These improvements stem from causal latent encoding and frame-wise sampling schedule of CMDM, which condition each frame on partially denoised preceding frames to ensure stability and temporal consistency. Please refer to the supplementary videos for the complete motion sequences and more visualization results. 

\subsection{Analysis}
\label{sec:efficiency}
\paragraph{Computational Efficiency}
Compared to other existing autoregressive methods, CMDM achieves a substantial improvement in inference efficiency through its advanced architecture and frame-wise sampling schedule, which significantly reduces generation time while maintaining superior motion realism. We evaluate the computational efficiency of our framework by generating 6-second motion sequences on an NVIDIA A100 GPU over 100 repetitions. As a result, MARDM operates with 310M parameters at 20~fps, and MotionStreamer with 318M parameters at 11~fps. In contrast, our proposed CMDM contains only 114M parameters (including both the MAC-VAE and Causal-DiT) and achieves 28~fps using the standard autoregressive process, and up to 125~fps with the proposed frame-wise sampling schedule. These highlight the remarkable efficiency of our causal diffusion framework and the effectiveness of the proposed sampling strategy for real-time motion generation.

\paragraph{Ablation Studies}
To investigate the impact of each component in CMDM, we conduct ablation studies on causal latent modeling, causal diffusion forcing, and the frame-wise sampling schedule (FSS) on HumanML3D.
As shown in~\Tref{tab:ablation}, replacing the MAC-VAE with a standard VAE significantly degrades motion quality, text fidelity, and transition smoothness, confirming the importance of causal latent modeling and semantic supervision.
Removing motion–language alignment (C-VAE w/o MA) produces motion quality comparable to MAC-VAE but introduces semantic inconsistencies, underscoring the importance of semantic features for maintaining fidelity and coherence.
Substituting causal diffusion with full-sequence diffusion (w/ Full-Seq. Diff.) increases transition FID and AUJ, verifying that per-frame causal diffusion enforces stronger temporal stability. Eliminating AdaLN (w/o AdaLN) or ROPE (w/o ROPE) results in higher FID and weaker long-horizon coherence, while removing both further amplifies these degradations.
Finally, the FSS variants show that smaller uncertainty scales ($L{=}5$) achieve smoother transitions and lower AUJ, whereas excessively large $L$ or smaller $K$ degrade stability. These results show that all the components jointly contribute to the high performance of CMDM.

\begin{table}[t]
\centering
\resizebox{\linewidth}{!}{
\begin{tabular}{l|cc|cc}
\toprule
\multirow{2}{*}{Methods}  & \multicolumn{2}{c|}{T2M Generation} & \multicolumn{2}{c}{Motion Transition} \\ 
\cline{2-5}
~ & R-Top1$\uparrow$ & FID$\downarrow$ & FID$\downarrow$ & AUJ$\downarrow$ \\ 
\midrule\midrule 
\rowcolor[gray]{0.90} CMDM & $\underline{0.588^{\pm.004}}$ & $\mathbf{0.068^{\pm.003}}$ & $\underline{1.66^{\pm0.06}}$ &  $\underline{0.42^{\pm0.01}}$\\ \midrule
VAE w/o MA & ${0.561^{\pm.005}}$ & ${0.107^{\pm.006}}$ & ${2.04^{\pm0.06}}$ &  ${0.52^{\pm0.01}}$ \\
C-VAE w/o MA & ${0.575^{\pm.004}}$ & ${0.070^{\pm.004}}$ & ${1.72^{\pm0.06}}$ &  ${0.44^{\pm0.01}}$ \\ \midrule
w/ Full-Seq. Diff.  & $\mathbf{0.591^{\pm.004}}$ & ${0.071^{\pm.003}}$ & ${1.96^{\pm0.07}}$ &  ${0.72^{\pm0.01}}$ \\ 
w/o adaLN & ${0.583^{\pm.002}}$ & ${0.076^{\pm.004}}$ & ${1.78^{\pm0.06}}$ &  ${0.47^{\pm0.01}}$ \\
w/o ROPE & ${0.581^{\pm.002}}$ & ${0.087^{\pm.004}}$  & ${1.85^{\pm0.08}}$ &  ${0.51^{\pm0.01}}$ \\
w/o adaLN + ROPE & ${0.580^{\pm.002}}$ & ${0.091^{\pm.005}}$ & ${2.33^{\pm0.07}}$ &  ${0.54^{\pm0.01}}$ \\
\midrule
w/o FSS  & ${0.576^{\pm.003}}$ & ${0.079^{\pm.003}}$ & ${1.82^{\pm0.05}}$ &  ${0.48^{\pm0.01}}$ \\ 
FSS $K=50, L=1$ & ${0.587^{\pm.005}}$ & $\underline{0.070^{\pm.003}}$ & ${1.71^{\pm0.05}}$ &  ${0.43^{\pm0.01}}$ \\
FSS $K=50, L=5$ & ${0.583^{\pm.005}}$ & ${0.077^{\pm.005}}$ & $\mathbf{1.64^{\pm0.06}}$ &  $\mathbf{0.38^{\pm0.01}}$\\
FSS $K=50, L=10$ & ${0.579^{\pm.004}}$ & ${0.080^{\pm.005}}$ & ${1.75^{\pm0.06}}$ &  ${0.45^{\pm0.01}}$ \\
FSS $K=20, L=1$ & ${0.585^{\pm.004}}$ & ${0.074^{\pm.004}}$  & ${1.70^{\pm0.05}}$ &  ${0.43^{\pm0.01}}$ \\
FSS $K=20, L=5$ & ${0.576^{\pm.005}}$ & ${0.088^{\pm.006}}$  & ${1.77^{\pm0.05}}$ &  ${0.42^{\pm0.01}}$  \\
FSS $K=20, L=10$ & ${0.572^{\pm.005}}$ & ${0.103^{\pm.005}}$ & ${1.75^{\pm0.05}}$ &  ${0.43^{\pm0.01}}$ \\
\bottomrule
\end{tabular}}
\vspace{-7pt}
\caption{Ablation studies of CMDM.}
\vspace{-11pt}
\label{tab:ablation}
\end{table}
\section{Limitations}
\label{sec:limitations}
Although CMDM achieves state-of-the-art performance in text-conditioned and long-horizon motion generation, several limitations remain. First, the causal latent encoding relies on motion–language alignment quality from pretrained motion-language models such as Part-TMR, which may limit performance when processing highly abstract or ambiguous text descriptions. Second, while the frame-wise sampling schedule substantially improves inference efficiency, it may still accumulate minor temporal artifacts when generating extremely long sequences, \eg, over several minutes. Incorporating motion-aware feedback or adaptive re-anchoring mechanisms could further improve long-horizon stability. Finally, CMDM focuses primarily on single-person motion and has not yet been extended to interactive or multi-character scenarios~\cite{tanaka2023interaction, liang2024intergen, fan2024freemotion, ota2025pino}, which will be an interesting direction for future work.
\section{Conclusion}
In this paper, we presented CMDM, a unified framework that combines the realism and stability of diffusion models with the temporal causality and efficiency of autoregressive generation. CMDM introduces a MAC-VAE for semantically grounded causal latent encoding, a Causal-DiT for temporally ordered diffusion denoising, and a FSS that enables real-time streaming generation. Extensive experiments on HumanML3D and SnapMoGen demonstrate that CMDM achieves superior motion fidelity, semantic alignment, and efficiency compared to existing diffusion and autoregressive models. We believe CMDM provides a promising step toward scalable, real-time, and semantically coherent motion generation.

{
    \small
    \bibliographystyle{ieeenat_fullname}
    \bibliography{main}
}

\appendix
\clearpage
\maketitlesupplementary

\section{Implementation Details}

\subsection{MAC-VAE}
The proposed MAC-VAE consists of seven causal convolutional layers and two causal ResNet blocks with left padding in both the encoder and decoder to ensure strict temporal causality. Each convolutional layer uses a kernel size of 3 and a stride of 1, followed by ReLU activation. The latent feature dimension is set to 64, and motion sequences are downsampled/upsampled by a factor of 4 along the temporal axis using stride-2 convolutional layers within the ResNet blocks.

To achieve semantic alignment between motion and text, we modify Part-TMR~\cite{yu2025remogpt} to extract frame-level motion–language embeddings. Part-TMR uses a \texttt{[class]} token to aggregate frames into a global feature, whereas we directly extract features from each frame and align them with the corresponding text features via contrastive learning, which serves as the supervision signal for MAC-VAE. The loss weighting coefficient is set to $\beta{=}1.0$, and the margin parameters are set to $m_1{=}0.5$ and $m_2{=}0.25$.

We train MAC-VAE using the AdamW optimizer with a learning rate of $1{\times}10^{-4}$ and a batch size of 128 for 50 epochs on a single NVIDIA A100 GPU. The learning rate follows a cosine decay schedule, and gradient clipping with a maximum norm of 1.0 is applied for training stability.

\subsection{Causal-DiT}
The Causal-DiT is implemented as a lightweight transformer-based denoiser with 8 layers, 4 attention heads, and a hidden dimension of 512. Causal self-attention is applied using a lower-triangular mask to enforce temporal order, while cross-attention conditions motion latents on text embeddings extracted from DistilBERT~\cite{sanh2019distilbert}. We incorporate Adaptive Layer Normalization (AdaLN)~\cite{peebles2023scalable} and Rotary Positional Encoding (ROPE)~\cite{su2024roformer} to embed timestep information and stabilize long-horizon attention. During training, the text condition is randomly dropped with a probability of 0.1 to enable classifier-free guidance. The model is optimized using AdamW with the same hyperparameter settings as MAC-VAE. The scale of classifier-free guidance is set to 3.0 during inference.

\subsection{Causal Diffusion Forcing}
In CMDM, causal diffusion forcing is employed to enable temporally ordered denoising while maintaining frame-level stochasticity. During training, each frame $t$ is perturbed with an independent noise level $k_t \in [0, K]$, where $K{=}1000$ denotes the total number of diffusion steps. The Causal-DiT serves as the denoiser, learning to predict noise residuals $\boldsymbol{\epsilon}_\theta(\tilde{\mathbf{z}}_{\le t}, k_t, \mathbf{c})$ conditioned on all preceding latent frames and the text embedding $\mathbf{c}$. This formulation ensures that each frame is denoised based solely on its causal history, thereby enforcing strict temporal dependencies. The overall training process is summarized in Algorithm~\ref{alg:cmdm-train}.

During inference, we adopt the Frame-Wise Sampling Schedule (FSS) with diffusion steps $K{=}50$ and uncertainty scale $L{=}2$. In this setting, the denoising of frame $t{+}1$ begins at step $K{-}L$ of frame $t$, allowing partially denoised frames to guide subsequent generations. This causal scheduling mechanism significantly accelerates inference by reducing redundant diffusion steps while maintaining temporal consistency across frames. The overall inference process with FSS is summarized in Algorithm~\ref{alg:cmdm-sample}.

\begin{algorithm*}[t]
\caption{CMDM Training with Causal Diffusion Forcing}
\label{alg:cmdm-train}
\begin{algorithmic}[1]
\Require Pretrained MAC\mbox{-}VAE encoder $E_\phi$, text embedding $\mathbf{c}$, Causal\mbox{-}DiT $\epsilon_\theta$, diffusion schedule $\{\alpha_k,\bar{\alpha}_k\}_{k=0}^K$
\For{\texttt{each minibatch}}
  \State Encode motion sequence: $\mathbf{z}_{1:T} \leftarrow E_\phi(\mathbf{x}_{1:T})$
  \For{$t = 1$ to $T$}
     \State Sample independent noise level $k_t \sim \mathcal{U}\{0, 1, \dots, K\}$
     \State Diffuse latent: $\tilde{\mathbf{z}}_t^{k_t} \leftarrow 
            \sqrt{\bar{\alpha}_{k_t}}\mathbf{z}_t + \sqrt{1-\bar{\alpha}_{k_t}}\boldsymbol{\epsilon}_t$,
            \quad $\boldsymbol{\epsilon}_t \sim \mathcal{N}(0,\mathbf{I})$
     \State Predict noise with causal conditioning: 
            $\hat{\boldsymbol{\epsilon}}_t \leftarrow 
            \epsilon_\theta(\tilde{\mathbf{z}}_{\le t}, k_t, \mathbf{c})$
  \EndFor
  \State Compute loss: 
        $\mathcal{L}_{\text{DF}} \leftarrow 
        \frac{1}{T}\sum_{t=1}^{T}\|\boldsymbol{\epsilon}_t - \hat{\boldsymbol{\epsilon}}_t\|_2^2$
  \State Update $\theta \leftarrow \theta - \eta\nabla_\theta \mathcal{L}_{\text{DF}}$
\EndFor
\end{algorithmic}
\end{algorithm*}

\begin{algorithm*}[t]
\caption{CMDM Streaming Generation with Frame-wise Sampling Schedule (FSS)}
\label{alg:cmdm-sample}
\begin{algorithmic}[1]
\Require Causal\mbox{-}DiT $\epsilon_\theta$, MAC\mbox{-}VAE decoder $D_\psi$, text embedding $\mathbf{c}$, schedule matrix $\mathbf{K}\in\mathbb{N}^{M\times T}$
\State Initialize $\tilde{\mathbf{z}}_t^{K} \sim \mathcal{N}(0,\mathbf{I})$ for $t{=}1,\dots,T$
\For{$m=1,\dots,M$}
  \For{$t=1,\dots,T$}
    \State Obtain noise level $k \leftarrow K_{m,t}$
    \State Predict noise with previous frames: $\hat{\boldsymbol{\epsilon}}_t \leftarrow \epsilon_\theta(\tilde{\mathbf{z}}_{\le t}^{\,k}, k, \mathbf{c})$
    \State 
      Denoise the current frame: $\tilde{\mathbf{z}}_t^{\,k-1} \leftarrow 
      \frac{1}{\sqrt{\alpha_k}}\!\left(\tilde{\mathbf{z}}_t^{\,k} - \frac{1-\alpha_k}{\sqrt{1-\bar{\alpha}_k}}\,\hat{\boldsymbol{\epsilon}}_t \right)
      + \sigma_k\,\mathbf{w},\ \mathbf{w}\!\sim\!\mathcal{N}(0,\mathbf{I})$
    \If{$k = 0$}
    \State Decode final clean latent: $\hat{\mathbf{x}}_t \leftarrow D_\psi(\tilde{\mathbf{z}}_{\le t}^{\,0})$
\EndIf
  \EndFor
\EndFor
\State \Return Decoded motion $\hat{\mathbf{x}}_{1:T}$ (or latents $\tilde{\mathbf{z}}_{1:T}^{\,0}$)
\end{algorithmic}
\end{algorithm*}

\section{Additional Quantitative Results}
\subsection{Experiments on BABEL}
We further evaluate CMDM on the BABEL dataset~\cite{punnakkal2021babel} to assess its generalization ability to diverse motion compositions. 
BABEL contains densely annotated sequences with multiple actions and transitions, making it suitable for long-horizon motion synthesis and evaluation.
We train CMDM by constructing training samples from adjacent subsequences in BABEL, where each pair of consecutive segments is used to learn motion continuation across long sequences.
As shown in~\Tref{tab:long_comparison_babel}, our method achieves the best overall performance across both subsequence and transition metrics, demonstrating the advantage of CMDM in maintaining consistency across action boundaries and generating smooth, continuous motions. 

\begin{table*}[t]
\centering
\resizebox{0.85\linewidth}{!}{
\begin{tabular}{l|cccc|cccc}
\toprule
\multirow{2}{*}{Methods}  & \multicolumn{4}{c|}{Subsequence} & \multicolumn{4}{c}{Transition} \\ \cline{2-9}
~ & R-prec$\uparrow$ & FID$\downarrow$ & Div$\rightarrow$ & MM-Dist$\downarrow$ & FID$\downarrow$ & Div$\rightarrow$ & PJ$\rightarrow$ & AUJ$\downarrow$ \\ 
\midrule \midrule 
GT & $0.715^{\pm0.003}$ & $0.00^{\pm0.00}$ & $8.42^{\pm0.15}$ & $3.36^{\pm0.06}$ & $0.00^{\pm0.00}$ & $6.20^{\pm0.06}$ & $0.02^{\pm0.00}$ & $0.00^{\pm0.00}$ \\ \midrule
FlowMDM & ${0.702^{\pm0.004}}$ & ${0.99^{\pm0.04}}$ & ${8.36^{\pm0.13}}$ & ${3.45^{\pm0.02}}$ & ${2.61^{\pm0.06}}$ & $\mathbf{6.47^{\pm0.05}}$ & ${0.06^{\pm0.00}}$ & ${0.13^{\pm0.00}}$ \\
\rowcolor[gray]{0.90} \textbf{Ours} & $\mathbf{0.711^{\pm0.005}}$ & $\mathbf{0.90^{\pm0.06}}$ & $\mathbf{8.47^{\pm0.20}}$ & $\mathbf{3.39^{\pm0.05}}$ & $\mathbf{2.45^{\pm0.05}}$ & ${6.73^{\pm0.05}}$ & $\mathbf{0.05^{\pm0.00}}$ & $\mathbf{0.11^{\pm0.01}}$ \\
\bottomrule
\end{tabular}}
\vspace{-7pt}
\caption{Comparison of long-horizon motion generation on BABEL. 
Subsequence metrics evaluate motion quality and diversity within segments, while transition metrics assess temporal continuity and smoothness between segments.}
\vspace{-7pt}
\label{tab:long_comparison_babel}
\end{table*}

\subsection{Evaluation on Other Motion Features}
To further examine the generalization ability of CMDM, we conduct experiments using motion features with redundant dimensions removed, following the analysis in~\cite{meng2024rethinking}. 
As discussed in prior work, the standard HumanML3D motion representation contains redundant components such as local joint rotations and contact features that do not directly influence the final human pose. 
Removing these redundant features yields a more compact and physically meaningful representation better suited for continuous diffusion modeling.

\Tref{tab:result_humanml3d_reduced} reports the results on HumanML3D using only essential motion features. 
Compared to the baseline methods, CMDM consistently improves generation quality and semantic alignment under both autoregressive (AR) and diffusion (FSS) configurations. 
Specifically, \textbf{CMDM w/ FSS} achieves the best overall performance, reaching an R-Precision of 0.563/0.759/0.849 for Top-1/Top-2/Top-3 accuracy and the lowest FID of 0.078, confirming that our causal diffusion formulation effectively models temporally coherent motion even in compact feature spaces. 
These results demonstrate that CMDM remains robust across different motion representations, further validating its adaptability to feature compression and reparameterized motion distributions.

\begin{table*}[t]
    \centering
    \resizebox{1\linewidth}{!}{
    \begin{tabular}{l| c | c c c| c |c| c| c}
    \toprule
      \multirow{2}{*}{Methods} & \multirow{2}{*}{Framework}  & \multicolumn{3}{c|}{R-Precision$\uparrow$} & \multirow{2}{*}{FID$\downarrow$} & \multirow{2}{*}{MM-Dist$\downarrow$} & \multirow{2}{*}{MModality$\uparrow$} & \multirow{2}{*}{CLIP-score$\uparrow$}\\
    \cline{3-5}
     ~ & ~& Top 1 & Top 2 & Top 3 & & & & \\
    \midrule\midrule
    T2M-GPT~\cite{zhang2023t2m} & \multirow{3}{*}{VQ}
    & $0.470^{\pm.003}$ & $0.659^{\pm.002}$ & $0.758^{\pm.002}$ 
    & $0.335^{\pm.003}$ & $3.505^{\pm.017}$ & $2.018^{\pm.053}$ & $0.607^{\pm.005}$ \\

    MMM~\cite{pinyoanuntapong2024mmm} &
    & $0.487^{\pm.003}$ & $0.683^{\pm.002}$ & $0.782^{\pm.002}$ 
    & $0.132^{\pm.004}$ & $3.359^{\pm.019}$ & $2.241^{\pm.073}$ & $0.635^{\pm.003}$ \\

    MoMask~\cite{guo2024momask} &
    & $0.490^{\pm.004}$ & $0.687^{\pm.003}$ & $0.786^{\pm.003}$ 
    & ${0.116^{\pm.006}}$ & $3.353^{\pm.010}$ & $1.263^{\pm.079}$ & ${0.637^{\pm.003}}$ \\
    \midrule

    MDM-50Step~\cite{mdm2022human} & \multirow{4}{*}{Diffusion}
    & $0.440^{\pm.007}$ & $0.636^{\pm.006}$ & $0.742^{\pm.004}$ 
    & $0.518^{\pm.032}$ & $3.640^{\pm.028}$ & $\mathbf{3.604^{\pm.031}}$ & $0.578^{\pm.003}$ \\

    MotionDiffuse~\cite{mdm2022human} &
    & $0.450^{\pm.006}$ & $0.641^{\pm.005}$ & $0.753^{\pm.005}$ 
    & $0.778^{\pm.035}$ & $3.490^{\pm.023}$ & $3.179^{\pm.046}$ & $0.606^{\pm.004}$ \\

    MLD~\cite{chen2023executing} &
    & $0.461^{\pm.004}$ & $0.651^{\pm.004}$ & $0.750^{\pm.003}$ 
    & $0.431^{\pm.014}$ & $3.445^{\pm.019}$ & $\underline{3.506^{\pm.031}}$ & $0.615^{\pm.003}$ \\

    ReMoDiffuse~\cite{zhang2023remodiffuse} &
    & $0.468^{\pm.003}$ & $0.653^{\pm.003}$ & $0.754^{\pm.005}$ 
    & $0.883^{\pm.021}$ & $3.414^{\pm.020}$ & $2.703^{\pm.154}$ & $0.621^{\pm.003}$ \\

    SALAD~\cite{hong2025salad} &
    & $\underline{0.552^{\pm.003}}$ & $\underline{0.748^{\pm.003}}$ & $\underline{0.839^{\pm.002}}$ 
    & $0.124^{\pm.005}$ & $2.990^{\pm.010}$ & $1.833^{\pm.081}$ & $0.671^{\pm.001}$ \\
    \midrule

    MARDM-DDPM~\cite{meng2024rethinking} & Autoregressive
    & ${0.492^{\pm.006}}$ & ${0.690^{\pm.005}}$ & ${0.790^{\pm.005}}$
    & $0.116^{\pm.004}$ & ${3.349^{\pm.010}}$ & ${2.470^{\pm.053}}$ & ${0.637^{\pm.005}}$ \\

    MARDM-SiT~\cite{meng2024rethinking} & Transformer
    & ${0.500^{\pm.004}}$ & ${0.695^{\pm.003}}$ & ${0.795^{\pm.003}}$
    & ${0.114^{\pm.007}}$ & ${3.270^{\pm.009}}$ & $2.231^{\pm.071}$ & ${0.642^{\pm.002}}$ \\ \midrule

    \textbf{Ours w/ AR} & Autoregressive
    & $0.550^{\pm.004}$ & $0.747^{\pm.004}$ & $0.838^{\pm.003}$
    & $\underline{0.085^{\pm.004}}$ & $\underline{2.987^{\pm.011}}$ & $1.810^{\pm.068}$ & $\underline{0.675^{\pm.001}}$ \\

    \textbf{Ours w/ FSS} & Diffusion
    & $\mathbf{0.563^{\pm.004}}$ & $\mathbf{0.759^{\pm.003}}$ & $\mathbf{0.849^{\pm.002}}$
    & $\mathbf{0.078^{\pm.003}}$ & $\mathbf{2.920^{\pm.007}}$ & $1.827^{\pm.094}$ & $\mathbf{0.685^{\pm.001}}$ \\
    \bottomrule
    \end{tabular}}
    \vspace{-7pt}
    \caption{Results of text-to-motion generation on HumanML3D without redundant features.
    The average is reported over 10 runs with 95\% confidence intervals.
    \textbf{Bold} indicates the best result, and \underline{underline} denotes the second-best result.}
    \vspace{-10pt}
    \label{tab:result_humanml3d_reduced}
\end{table*}

\subsection{Compositional Motion Generation}
\label{sec:compositional}

We evaluate CMDM on the compositional motion generation task following the protocol of Multi-Track Timeline (MTT)~\cite{petrovich2024multi}, which requires generating coherent motions conditioned on multiple temporally structured text descriptions. This task evaluates both semantic composition, \ie, correctly realizing multiple concepts within a single sequence, and temporal composition, \ie, ensuring smooth and consistent transitions across segments.

Specifically, following prior work~\cite{petrovich2024multi, zhang2025energymogen}, we report per-crop semantic correctness metrics (R@1, R@3, and TMR-Score for M2T and M2M), as well as realism metrics including FID and transition distance. As shown in Table~\ref{tab:compositional_mtt}, CMDM, under the single-track multi-crop setting, consistently outperforms EnergyMoGen and other compositional baselines across all metrics. Notably, CMDM achieves substantial improvements in semantic alignment while simultaneously reducing FID and transition distance, demonstrating stronger long-horizon consistency and smoother transitions between composed motion segments.

\begin{table}[t]
\centering
\renewcommand{\arraystretch}{0.9}
\setlength{\tabcolsep}{1pt}
\resizebox{\linewidth}{!}{
\small
\begin{tabular}{l|cc|cc|cc|cc}
\toprule
\multirow{2}{*}{Method}
& \multicolumn{2}{c|}{Input type}
& \multicolumn{4}{c|}{Per-crop semantic correctness}
& \multicolumn{2}{c}{Realism} \\
& \#tracks & \#crops
& R@1$\uparrow$ & R@3$\uparrow$
& M2T$\uparrow$ & M2M$\uparrow$
& FID$\downarrow$ & Transition$\downarrow$ \\
\midrule

GT 
& - & - 
& $55.0$ & $73.3$ & $0.748$ & $1.000$ 
& $0.000$ & $1.5$ \\
\midrule

EnergyMoGen~\cite{zhang2025energymogen} 
& Single & Single 
& $15.9$ & $28.0$ & $0.591$ & $0.567$ 
& $0.604$ & $1.6$ \\

MDM-SMPL~\cite{petrovich2024multi} 
& Single & Single 
& $12.1$ & $23.5$ & $0.573$ & $0.578$ 
& $0.484$ & $1.8$ \\

w/ DiffCollage~\cite{petrovich2024multi} 
& Single & Multi 
& $29.1$ & $49.7$ & $0.675$ & $0.656$ 
& $0.446$ & $1.2$ \\

w/ STMC~\cite{petrovich2024multi} 
& Multi & Multi 
& $30.5$ & $50.9$ & $0.675$ & $0.665$ 
& $0.459$ & $\mathbf{0.9}$ \\

\rowcolor[gray]{0.90}
\textbf{CMDM (Ours)}
& Single & Multi 
& $\mathbf{41.7}$ & $\mathbf{57.9}$ 
& $\mathbf{0.690}$ & $\mathbf{0.672}$ 
& $\mathbf{0.438}$ & $1.2$ \\

\bottomrule
\end{tabular}
}
\vspace{-8pt}
\caption{
Comparison with prior compositional motion generation methods on the Multi-track timeline (MTT) dataset~\cite{petrovich2024multi}. 
}
\vspace{-8pt}
\label{tab:compositional_mtt}
\end{table}

\subsection{Latency analysis}
To evaluate the practical efficiency of different causal motion generation methods, we measure the latency for generating each token (4 frames) on a single NVIDIA A100 GPU.
MARDM~\cite{meng2024rethinking}, MotionStreamer~\cite{xiao2025motionstreamer}, and CMDM w/ AR require approximately 210\,ms, 360\,ms, and 150\,ms, respectively, to generate the first token, with similar latency for each subsequent token.
This is because these autoregressive diffusion methods perform full diffusion denoising for each token independently, requiring multiple denoising steps per frame regardless of its temporal position.
In contrast, CMDM w/ FSS takes about 220\,ms for the first token but only 30\,ms per subsequent token, achieving a $5\times$--$12\times$ speedup for streaming generation.
This dramatic reduction in per-token latency stems from our frame-wise sampling schedule, which allows each frame to be predicted from partially denoised preceding frames rather than requiring full iterative refinement.

\subsection{Ablation Studies}
\paragraph{Architecture of MAC-VAE.}
We evaluate several configurations of MAC-VAE to analyze the effects of latent dimension and temporal downsampling rate on both reconstruction and generation performance. 
The notation $(d, r)$ denotes the latent dimension $d$ and the temporal downsampling rate $r$. 
As shown in ~\Tref{tab:result_vae_ab}, increasing the latent dimension improves reconstruction accuracy but also introduces redundancy that slightly affects generation quality in terms of FID. 
Conversely, larger temporal downsampling rates (\eg, $r=1/8$) reduce temporal resolution and lead to minor degradation in R-Precision and MM-Dist due to information loss. 
Among all configurations, MAC-VAE with $(64, 1/4)$ achieves the best balance between reconstruction fidelity (FID$=0.000$, MPJPE$=0.012$) and generation quality (R-Top1$=0.588$, FID$=0.068$, MM-Dist$=2.620$), which we adopt as the default setting in all subsequent experiments. 
These results confirm that a compact latent space with moderate temporal compression effectively captures semantic and temporal dependencies for downstream motion generation.

\begin{table}[t]
    \centering
    \resizebox{1\linewidth}{!}{
    \begin{tabular}{l | c | c c | c c c }
    \toprule
      \multirow{2}{*}{Model} & \multirow{2}{*}{Config.}  
      & \multicolumn{2}{c|}{Reconstruction}   
      & \multicolumn{3}{c}{Generation} \\ 
      \cline{3-7}
      & & {FID$\downarrow$} & {MPJPE$\downarrow$}  
      & {R-Top1$\uparrow$} & {FID$\downarrow$} & {MM-Dist$\downarrow$} \\ 
      \midrule\midrule
      VAE & 64, 1/4 & $0.001$ & $0.016$ & $0.561$ & $0.107$ & $2.706$ \\ 
      C-VAE & 64, 1/4 & $0.000$  & $0.012$ & $0.575$ & $0.070$ & $2.650$ \\ 
      \midrule
      \rowcolor[gray]{0.90} MAC-VAE & 64, 1/4 & $\mathbf{0.000}$ & $\mathbf{0.012}$ & $\mathbf{0.588}$ & $0.068$ & $\mathbf{2.620}$ \\
      MAC-VAE & 32, 1/4 & $0.002$ & $0.033$ & $0.583$ & $0.065$ & $2.628$ \\
      MAC-VAE & 16, 1/4 & $0.011$ & $0.077$ & $0.573$ & $0.071$ & $2.647$ \\ \midrule
      MAC-VAE & 64, 1/8 & $0.002$ & $0.035$ & $0.570$ & $0.069$ & $2.664$ \\
      MAC-VAE & 32, 1/8 & $0.006$ & $0.060$ & $0.566$ & $0.057$ & $2.704$ \\
      MAC-VAE & 16, 1/8 & $0.024$ & $0.101$ & $0.561$ & $\mathbf{0.054}$ & $2.709$ \\
    \bottomrule
    \end{tabular}}
    \vspace{-8pt}
    \caption{Comparison of reconstruction and generation performance on HumanML3D.  MPJPE is measured in millimeters. The notation $(d, r)$ denotes the latent dimension $d$ and the temporal downsampling rate $r$.}
    \vspace{-8pt}
    \label{tab:result_vae_ab}
\end{table}

\paragraph{Motion-Language Models}
To evaluate the effectiveness of different motion–language alignment strategies, we compare several pretrained motion–language models integrated into the MAC-VAE framework, including TMR~\cite{petrovich2023tmr}, MotionPatches~\cite{yu2024exploring}, and Part-TMR~\cite{yu2025remogpt}. 
As shown in~\Tref{tab:result_mlm_ab}, all motion–language models improve generation quality while maintaining reconstruction performance compared to the baseline VAE and C-VAE, demonstrating the effectiveness of semantic alignment between motion and text.
Among them, Part-TMR achieves the best overall performance with the lowest reconstruction error (FID$=0.000$, MPJPE$=0.012$) and the highest R-Precision ($0.588$), confirming its strong ability to capture fine-grained part-level correspondences between text and motion. 
These results validate the choice of Part-TMR as the alignment backbone in MAC-VAE, enabling more semantically coherent and temporally consistent motion generation.

\begin{table}[t]
    \centering
    \resizebox{1\linewidth}{!}{
    \begin{tabular}{l | c c | c c c }
    \toprule
      \multirow{2}{*}{Model} 
      & \multicolumn{2}{c|}{Reconstruction}   
      & \multicolumn{3}{c}{Generation} \\ 
      \cline{2-6}
      & {FID$\downarrow$} & {MPJPE$\downarrow$}  
      & {R-Top1$\uparrow$} & {FID$\downarrow$} & {MM-Dist$\downarrow$} \\ 
      \midrule\midrule
      VAE & $0.001$ & $0.016$ & $0.561$ & $0.107$ & $2.706$ \\ 
      C-VAE & $0.001$  & $0.012$ & $0.575$ & $0.070$ & $2.650$ \\
      \midrule
      \rowcolor[gray]{0.90} Part-TMR~\cite{yu2025remogpt} & $\mathbf{0.000}$ & $\mathbf{0.012}$ & $\mathbf{0.588}$ & $\mathbf{0.068}$ & $\mathbf{2.620}$ \\
      MotionPatches~\cite{yu2024exploring} & $0.000$ & $0.013$ & $0.586$ & $0.070$ & $2.622$ \\
      TMR~\cite{petrovich2023tmr} & $0.001$ & $0.013$ & $0.580$ & $0.070$ & $2.638$ \\ 
    \bottomrule
    \end{tabular}}
    \vspace{-6pt}
    \caption{Comparison of motion-language models in MAC-VAE on HumanML3D. MPJPE is measured in millimeters.}
    \vspace{-6pt}
    \label{tab:result_mlm_ab}
\end{table}

\paragraph{Model Size of Causal-DiT.}
We investigate the impact of model size on generation quality by varying the number of attention heads $(H)$, layers $(L)$, and hidden dimensions $(D)$ in Causal-DiT. 
As shown in~\Tref{tab:result_dit_ab}, larger models generally achieve better performance due to increased representational capacity. 
The medium-sized model (38M parameters) already provides strong results with an R-Precision of $0.588$ and FID of $0.068$, balancing quality and efficiency. 
Further scaling to 304M parameters yields marginal improvements (R-Precision$=0.590$, FID$=0.042$), demonstrating that Causal-DiT scales effectively while maintaining computational practicality. 
Unless otherwise specified, we use the medium (38M) configuration in all main experiments.

\begin{table}[t]
    \centering
    \resizebox{1\linewidth}{!}{
    \begin{tabular}{l | c | c c c | c c }
    \toprule
      \multirow{2}{*}{Model Size} & \multirow{2}{*}{Config.}  
      & \multicolumn{3}{c|}{R-Precision$\uparrow$}   
      & \multirow{2}{*}{FID$\downarrow$} & \multirow{2}{*}{MM-Dist$\downarrow$} \\ 
      \cline{3-5}
      & & {Top1} & {Top2} & {Top3} \\ 
      \midrule\midrule
       S (19M) & H2, L4, D512 & $0.543$ & $0.738$ & $0.834$ & $0.247$ & $2.845$ \\
      \rowcolor[gray]{0.90} M (38M) & H4, L8, D512 & $0.588$ & $0.778$ & $0.860$ & $0.068$ & $2.620$ \\
      L (129M) & H6, L12, D768 & $0.585$ & $0.779$ & $0.859$ & $0.044$ & $2.621$ \\
      XL (304M) & H8, L16, D1024 & $\mathbf{0.590}$ & $\mathbf{0.779}$ & $\mathbf{0.861}$ & $\mathbf{0.042}$ & $\mathbf{2.610}$ \\
    \bottomrule
    \end{tabular}}
    \vspace{-6pt}
    \caption{Comparison of model sizes on HumanML3D. The notation $(H, L, D)$ denotes the number of attention heads $H$, layers $L$, and hidden dimension $D$.}
    \vspace{-6pt}
    \label{tab:result_dit_ab}
\end{table}

\paragraph{Text Encoder}

\begin{table}[t]
    \centering
    \resizebox{1\linewidth}{!}{
    \begin{tabular}{l | c | c c c | c c }
    \toprule
      \multirow{2}{*}{Text Encoder}  & \multirow{2}{*}{Embedding}  
      & \multicolumn{3}{c|}{R-Precision$\uparrow$}   
      & \multirow{2}{*}{FID$\downarrow$} & \multirow{2}{*}{MM-Dist$\downarrow$} \\ 
      \cline{3-5}
      & & {Top1} & {Top2} & {Top3} \\ 
      \midrule\midrule
      \rowcolor[gray]{0.90} DistilBERT~\cite{sanh2019distilbert} & Word & $\mathbf{0.588}$ & $\mathbf{0.778}$ & $\mathbf{0.860}$ & $\mathbf{0.068}$ & $\mathbf{2.620}$ \\
      CLIP~\cite{radford2021learning} & Word &  $0.556$ & $0.751$ & $0.843$ & $0.086$ & $2.717$ \\
      CLIP~\cite{radford2021learning} & Sentence &  $0.527$ & $0.717$ & $0.809$ & $0.145$ & $2.941$ \\
      Sentence-T5~\cite{ni2022sentence} & Sentence &  $0.564$ & $0.754$ & $0.841$ & $0.126$ & $2.737$ \\
    \bottomrule
    \end{tabular}}
    \vspace{-6pt}
    \caption{Comparison of text encoders on HumanML3D.}
    \vspace{-6pt}
    \label{tab:result_text_ab}
\end{table}

\begin{figure*}[t]
    \centering
    \includegraphics[width=1\linewidth]{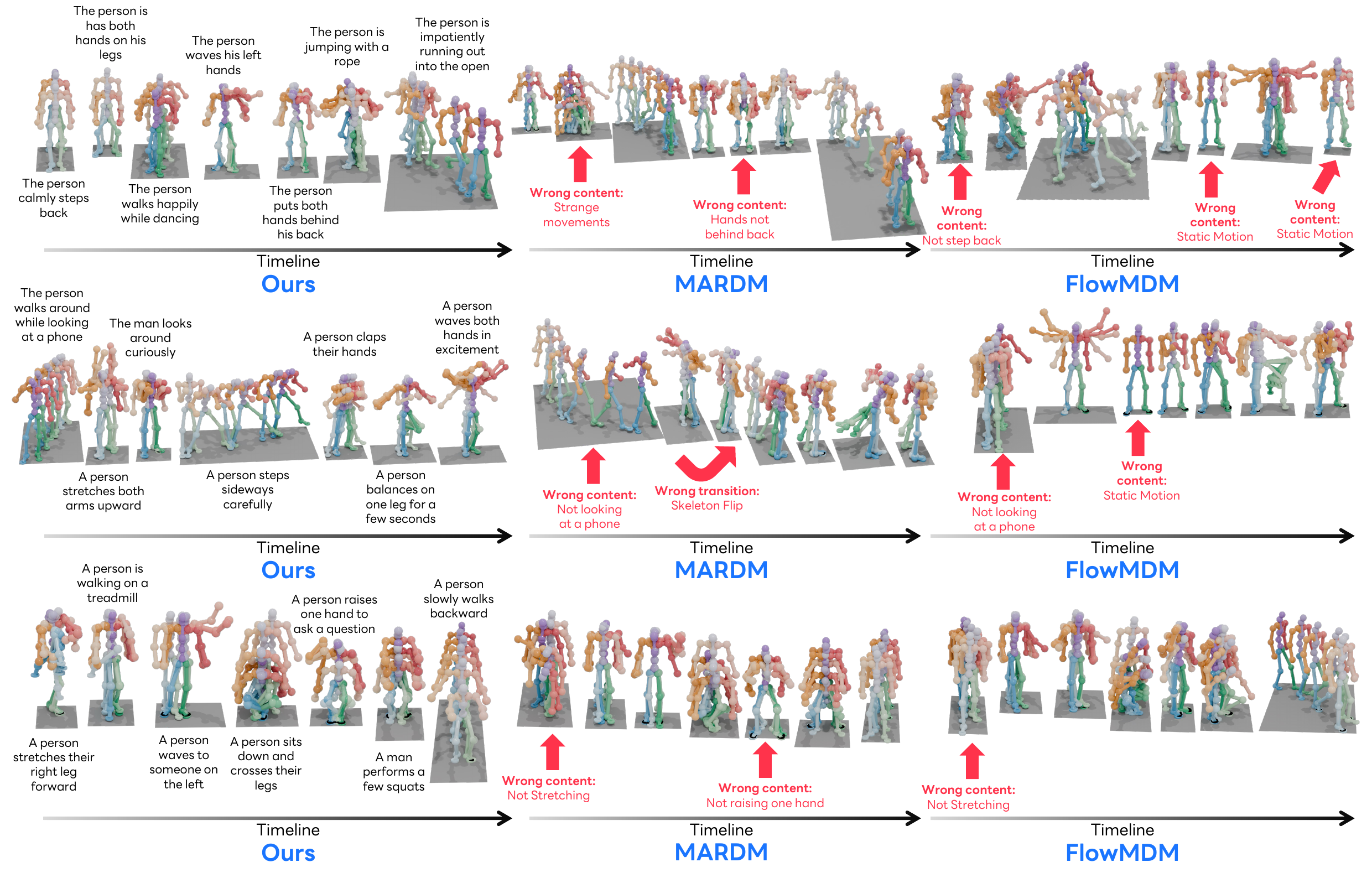}
    \vspace{-5pt}
    \caption{Qualitative results of long-horizon motion generation on HumanML3D. Comparison between our CMDM and previous methods. The generated motion is continuous and seamless; for visualization purposes, we split each long sequence into shorter segments corresponding to their captions. Please refer to the videos in the supplementary materials for the complete motion sequences.}
    \vspace{-5pt}
    \label{fig:qualitative_long_2}
\end{figure*}

\begin{figure*}[t]
    \centering
    \includegraphics[width=1\linewidth]{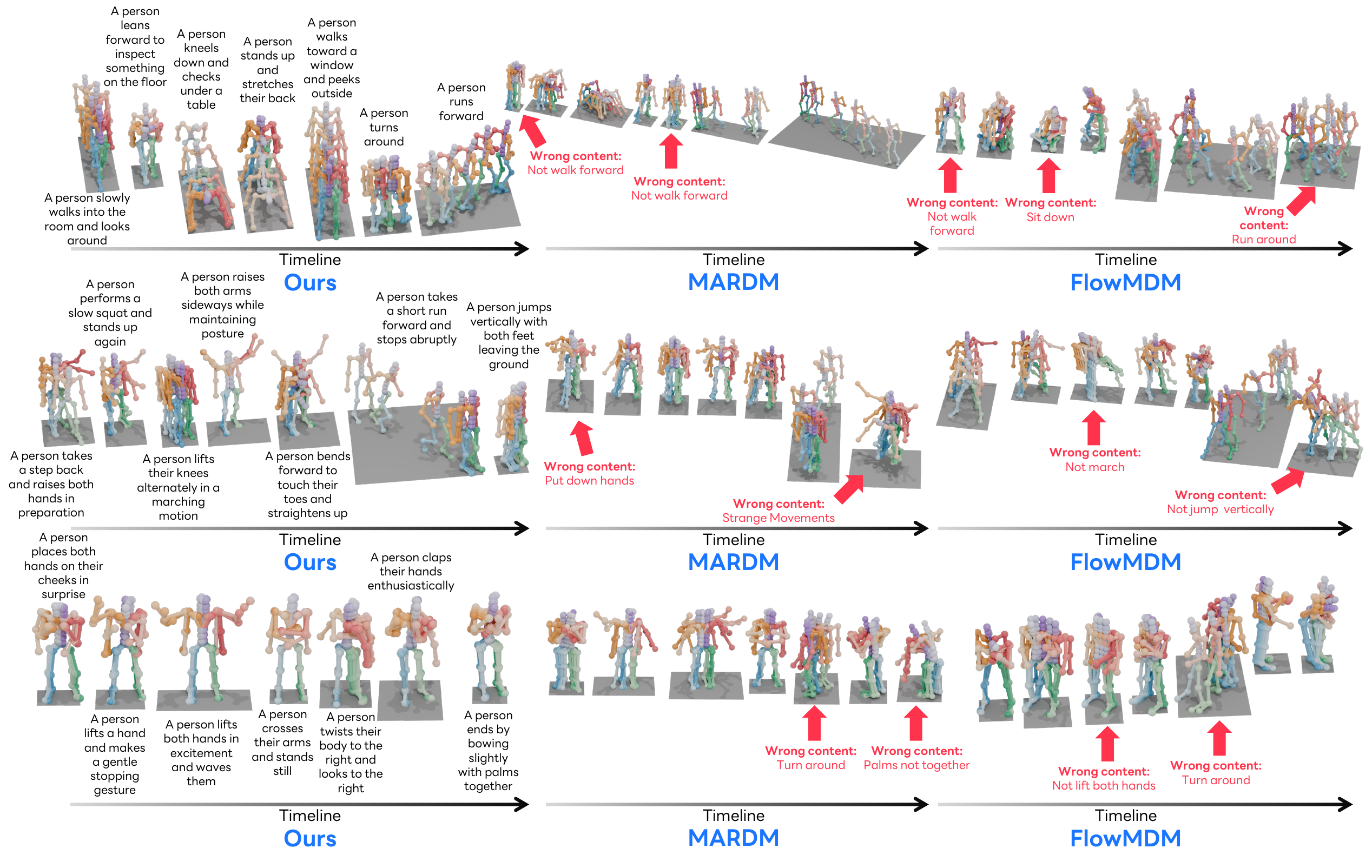}
    \vspace{-5pt}
    \caption{Qualitative results of long-horizon motion generation on SnapMoGen. Comparison between our CMDM and previous methods. The generated motion is continuous and seamless; for visualization purposes, we split each long sequence into shorter segments corresponding to their captions. Please refer to the videos in the supplementary materials for the complete motion sequences.}
    \vspace{-5pt}
    \label{fig:qualitative_long_snap}
\end{figure*}

We compare several pretrained language models as text encoders to evaluate their impact on semantic alignment and motion quality. As shown in~\Tref{tab:result_text_ab}, the choice of text encoder influences both text–motion correspondence (R-Precision) and visual realism (FID). DistilBERT~\cite{sanh2019distilbert}, which provides word-level embeddings, achieves the best overall performance with the highest R-Precision ($0.588$) and lowest FID ($0.068$), demonstrating its ability to capture fine-grained semantic cues that align well with motion features. Using the CLIP-based encoder, the word-level variant, which is identical to that employed in StableMoFusion~\cite{huang2024stablemofusion}, also outperforms StableMoFusion, further confirming the benefits of word-level representations. Such token-level embeddings are crucial for maintaining causal dependencies between linguistic tokens and motion frames, which is necessary for stable autoregressive generation in CMDM. In contrast, sentence-level embeddings from CLIP~\cite{radford2021learning} exhibit reduced precision and higher FID due to the loss of temporal granularity. Meanwhile, Sentence-T5~\cite{ni2022sentence} performs better than the CLIP-based models and also outperforms MotionLCM V2~\cite{huang2024stablemofusion}, despite MotionLCM V2 also using Sentence-T5. These findings validate our choice of DistilBERT as the text encoder for CMDM, as it effectively preserves local semantics and enables causally consistent motion–language modeling.

\begin{figure*}[t]
    \centering
    \includegraphics[width=\linewidth]{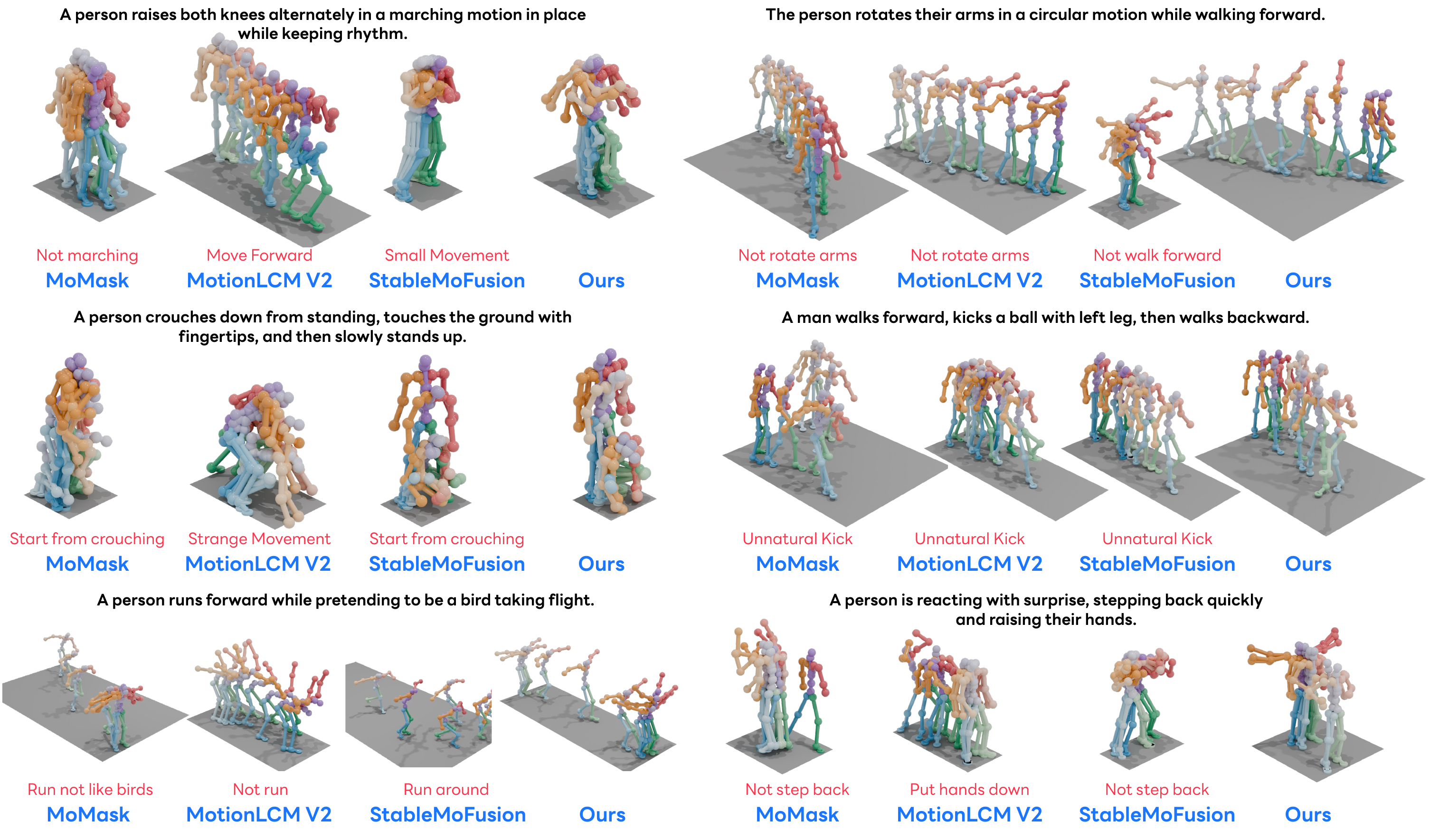}
    \vspace{-20pt}
    \caption{Qualitative results of text-to-motion generation on HumanML3D. CMDM produces motions that better capture fine-grained textual semantics and maintain natural body articulation compared to previous methods. Please refer to the supplementary videos for clearer visualization.}
    \vspace{-5pt}
    \label{fig:qualitative_t2m}
\end{figure*}

\begin{figure*}[t]
    \centering
    \includegraphics[width=\linewidth]{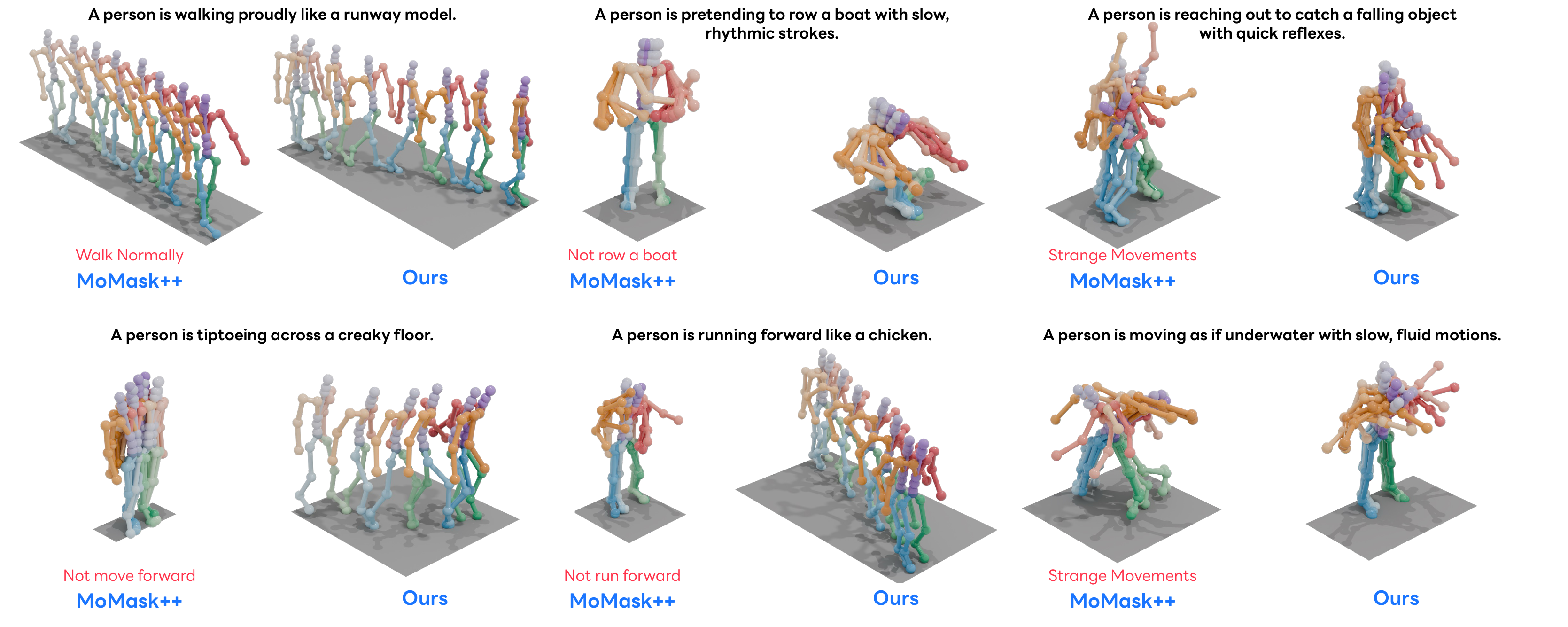}
    \vspace{-20pt}
    \caption{Qualitative results of text-to-motion generation on SnapMoGen. Comparison between our CMDM and previous methods. We directly use the raw text prompts without any LLM-based augmentation and CMDM still achieves strong generation quality. Please refer to the supplementary videos for clearer visualization.}
    \vspace{-5pt}
    \label{fig:qualitative_t2m_snap}
\end{figure*}

\section{Additional Qualitative results}
To further demonstrate the effectiveness of CMDM, we provide additional qualitative comparisons on long-horizon and text-to-motion generation. 
\Fref{fig:qualitative_long_2} and~\Fref{fig:qualitative_long_snap} compares CMDM with FlowMDM~\cite{barquero2024seamless} and MARDM~\cite{meng2024rethinking} on long-horizon motion generation for HumanML3D and SnapMoGen, respectively. 
CMDM produces temporally coherent and semantically accurate motions without content drift or skeleton flipping, whereas previous methods often suffer from static poses, incorrect transitions, or inconsistent actions across segments. 
These examples highlight the ability of CMDM to maintain smooth temporal dynamics and causal consistency throughout extended sequences.

\Fref{fig:qualitative_t2m} presents qualitative results on HumanML3D. 
Compared with MoMask~\cite{guo2024momask}, MotionLCM~\cite{dai2024motionlcm}, and StableMoFusion~\cite{huang2024stablemofusion}, CMDM generates motions that more faithfully reflect fine-grained textual semantics (\eg, arm rotations, leg movements, or walking direction) while preserving natural body articulation. 
\Fref{fig:qualitative_t2m_snap} shows additional results on SnapMoGen, where CMDM directly uses the raw text prompts without LLM-based augmentation and still produces more realistic motions than prior methods. 

\textbf{\textit{Please refer to the supplementary videos on the demo page for full-length visualizations.}}

\section{Sample Code}
The code will be released at \url{https://github.com/YU1ut/CMDM}.  We provide the training codes for building and evaluating the proposed CMDM with the HumanML3D dataset. \textbf{\textit{Please refer to the README file in the code directory for details.}}

\end{document}